%% file: main.tex
\documentclass[11pt,table]{article}

\pdfoutput=1

\usepackage[utf8]{inputenc} 
\usepackage[T1]{fontenc}    
\usepackage[in]{fullpage}
\usepackage{graphicx}
\usepackage{amsmath}
\usepackage{amssymb}
\usepackage{booktabs}
\usepackage[pagebackref,breaklinks,colorlinks]{hyperref}
\usepackage{url}            
\usepackage{booktabs}       
\usepackage{amsfonts}       
\usepackage{nicefrac}       
\usepackage{microtype}      
\usepackage{listings} 
\usepackage{subcaption}
\usepackage{multirow}
\usepackage{hhline}
\usepackage{colortbl}
\usepackage{amsmath}
\usepackage{xcolor}
\usepackage{textcomp}
\usepackage[textsize=scriptsize]{todonotes}      
\setuptodonotes{inline}

\definecolor{light-light-gray}{gray}{0.90} 

\usepackage[textsize=scriptsize]{todonotes}      

\usepackage[capitalize]{cleveref}
\crefname{section}{Sec.}{Secs.}
\Crefname{section}{Section}{Sections}
\Crefname{table}{Table}{Tables}
\crefname{table}{Tab.}{Tabs.}

\makeatletter
\renewcommand\paragraph{\@startsection{paragraph}{4}{\z@}                                     {1.05ex \@plus1ex \@minus.2ex}                                {-.5em}
{\normalfont\normalsize\bfseries}}
\makeatother

\usepackage[binary-units=true]{siunitx}
\DeclareSIUnit\flop{FLOP}
\DeclareSIUnit[per-mode=symbol]\floppersec{\flop\per\second}

\title{A Comparative Study on Generative Models for High Resolution Solar Observation Imaging}  

\author{
\textbf{Mehdi Cherti}$^{1}$  \quad \textbf{Alexander Czernik}$^{1,2}$  \quad \textbf{Stefan Kesselheim}$^{1,2}$   \\ \textbf{Frederic Effenberger}$^{3}$  \quad \textbf{Jenia Jitsev}$^{1}$  \\
Juelich Supercomputing Center (JSC), Research Center Juelich (FZJ)$^{1}$  \\ 
Helmholtz AI$^{2}$ \\
Ruhr-Universität Bochum$^{3}$
}

\begin{document}

\date{}
\maketitle
\begin{abstract}
Solar activity is one of the main drivers of variability in our solar system and the key source of space weather phenomena that affect Earth and near Earth space. The extensive record of high resolution extreme ultraviolet (EUV) observations from the Solar Dynamics Observatory (SDO) offers an unprecedented, very large dataset of solar images. In this work, we make use of this comprehensive dataset to investigate capabilities of current state-of-the-art generative models to accurately capture the data distribution behind the observed solar activity states. Starting from StyleGAN-based methods, we uncover severe deficits of this model family in handling fine-scale details of solar images when training on high resolution samples, contrary to training on natural face images. When switching to the diffusion based generative model family, we observe strong improvements of fine-scale detail generation. For the GAN family, we are able to achieve similar improvements in fine-scale generation when turning to ProjectedGANs, which uses multi-scale discriminators with a pre-trained frozen feature extractor. We conduct ablation studies to clarify mechanisms responsible for proper fine-scale handling. Using distributed training on supercomputers, we are able to train generative models for up to 1024x1024 resolution that produce high quality samples indistinguishable to human experts, as suggested by the evaluation we conduct. We make all code, models and workflows used in this study publicly available at \url{https://github.com/SLAMPAI/generative-models-for-highres-solar-images}.
\end{abstract}

\nocite{langley00}

\section{Introduction}
Generative models for high resolution images have seen a rapid progress in the last years, enabling for instance generation of highly photo-realistic, diverse natural image samples after training on large-scale data \cite{sauer2022stylegan,dhariwal2021diffusion,rombach2022high}. Consequently, different domains that operate on image-like signals were seeking to apply the powerful data-driven model class to study various domain-specific problems. Here, we pose the question whether in the domain of solar physics that offers high volume, high quality data on the state and dynamics of the sun recorded during long term observation missions, generative models can learn the underlying data distribution with a degree of realism sufficient for scientific requirements.

For this study, we take the large-scale dataset provided by the Solar Dynamic Observatory \cite{pesnell2012solar}, operated by NASA since 2010. Motivated by the success on high resolution natural image generation, we first focus on generative adversarial networks, GANs~\cite{goodfellow2014generative}, that show impressive results, producing for instance face and other images that are hard to distinguish from originals by human observers. Surprisingly, we discover that state-of-the-art GANs have troubles when it comes to faithful reproduction of important fine scale details of solar images even after careful tuning of the learning procedure, which is contrary to high quality fine scale generation for such natural images as faces. 

We then turn to another type of generative models - diffusion models. Using standard ablated diffusion models, ADM~\cite{dhariwal2021diffusion}, we are able to generate high resolution solar image samples that show sufficient quality of fine scale details. In order to match this capability with GANs, we experiment with ProjectedGAN~\cite{sauer2021projected}, a model type that introduces additional mechanisms to deal with multi-scale nature of images, mixing features and employing multiple discriminators across different scales. We observe that this GAN type can handle fine scale of solar images properly, producing samples comparable to diffusion models. We conduct ablation studies to confirm the importance of multi-scale mechanisms employed in ProjectedGAN for successfull fine-scale handling. 

To further assess the quality of generated samples, we conduct a small study with human observers, with results suggesting that it is impossible even for the experts from the solar imaging community to tell the generated from real sample images. We discuss the implications of our study and conclude that after training on a large volume of high quality scientific data, generative models are capable of producing realistic high resolution solar images with a level of detail that makes generated samples indistinguishable from real observations for human experts in the field.

We open-source the code for training, evaluating and the workflows around the dataset pre-processing. We hope that providing this to the community will result in follow-up research using high resolution - up to 4096x4096 - SDO images to further study generative modelling on an impacting scientific dataset that has high demands with regard to accurate generation of important details across different scales. 




\section{Related Work}

\textbf{Generative adversarial networks.}
Since the suggestion of GANs in 2014 \cite{goodfellow2014generative}, this approach has led to the generation of synthetic images of ever increasing quality in different domains. GANs seek to find a deep neural network $G\left(z, \theta_G \right)$ parametrized by weights $\theta_G$ that maps random numbers $z$ to generated images $x_g$, the generator. The antagonist, the discriminator $D$, parametrized by $\theta_D$, acts as a critic that is trained to distinguish the generated images from real images $x_r$ by estimating a probability the image is real. To aim for replicating the real data density, the generator should move along those directions in $\theta_G$-space which make the correct estimation for discriminator more difficult. A key ingredient is thus a formulation of loss $\mathcal L$ that generates a suitable gradient signal for both components. Formally, such loss to be minimized can be derived from a minimax problem:
\begin{align*}
\mathop{\text{min}}_{\theta_G} \mathop{\text{max}}_{\theta_D}  &  \mathop{\mathbb{E}}_{x \sim p_{r}(x)} \left[ \mathcal L_r \left( D\left( x; \theta_D \right) \right)  \right] + \\
&  \mathop{\mathbb{E}}_{z \sim p_z(z)} \left[ \mathcal L_g \left( D\left( G\left( z; \theta_G \right); \theta_D \right) \right) \right].
\end{align*}

The exact functional form of the loss, just as the family of neural networks of both generator and discriminator are under active research, and many extensions of this scheme e.g.~with regularizing terms have been proposed. While very successful, the minimax structure of the training objective makes the training notoriously unstable, leading to problems like mode collapse and non-convergence. 

Recently, GANs from the StyleGAN-family have been especially successful at generating images \cite{karras2019style}, where a key structure of the generator network is inspired by style transfer applications: before feeding the random numbers z into a standard generator, a multilayer perceptron preprocesses z and feeds its output into different layers of the generator. In StyleGAN2, the simple feedforward design of both the generator and the discriminator architecture are changed to skip networks and residual networks respectively~\cite{karras2020analyzing}. To enable data-efficient training and to overcome inherent instabilities leading to mode collapse, approaches like DiffAug \cite{zhao2020differentiable} and StyleGAN2-ADA\cite{karras2020training} transform images with differentiable augmentations before feeding them into the discriminator. This can lead to impressive results even under low data condition \cite{karras2020training}. StyleGAN3 eliminates texture sticking artifacts by employing signal processing techniques like Fourier feature mapping into the generator's architecture  \cite{karras2021alias}. Significant architectural changes to the discrimininator side were introduced with ProjectedGAN \cite{sauer2021projected}. ProjectedGAN introduces three main novelties there: (a) a frozen pretrained feature network is used to extract features from both generated and real images, (b) these features are embedded into a higher-dimensional space and mixed across scales, and (c) then fed into multiple trainable discriminators. ProjectedGAN reduces computational effort by introducing a frozen pre-trained feature network into which generated and real images are fed. Additionally, outputs obtained from different layers of the feature network are mixed before each output is processed by an independent discriminator \cite{sauer2021projected}. StyleGAN-XL \cite{sauer2022stylegan} scales this approach up by introducing progressive growing and leveraging labeled data through classifier guidance and pre-trained class embedding. With a StyleGAN3 generator increased in width, the authors generate images with impressive quality in a dataset with very diverse image content such as ImageNet \cite{deng2009imagenet}.



\textbf{Diffusion models.} Diffusion models~\cite{sohl2015deep,song2019generative,ho2020denoising} are a class of generative models that have shown impressive results in generative modelling, rivaling or surpassing GANs in terms of quality, mode coverage~\cite{theis2015note}, and diversity of the samples~\cite{dhariwal2021diffusion} and are the basis of recent breakthroughs in the text-to-image task in works such as DALL-E 2~\cite{ramesh2022hierarchical} and Imagen~\cite{saharia2022photorealistic}. Compared to GANs, they do not suffer from training instabilities and like other likelihood-based models, they provide better mode coverage~\cite{theis2015note}. The main downside of diffusion models is that they are much slower at generating samples (inference time) than GANs as they require multiple (denoising) steps, which can vary from hundreds to thousands of forward passes of the model. Following the original works~\cite{sohl2015deep,song2019generative,ho2020denoising}, several works have been proposed to make sample generation faster  by using a more efficient sampling procedure~\cite{song2020denoising,liu2022pseudo} or by estimating the denoising function with a multi-modal distribution~\cite{xiao2021tackling} or by using distillation~\cite{salimans2022progressive,meng2022distillation}. In another line of work, latent diffusion models~\cite{rombach2022high} (LDM) proposed to apply a denoising diffusion  model on a pre-trained \emph{VQGAN}~\cite{esser2021taming} latent space instead of raw pixels, focusing on high level details and leaving modelling fine-scale details to \emph{VQGAN}. LDMs have been shown to require much less denoising steps than vanilla diffusion models, and are the basis of the recent \emph{Stable Diffusion} that was used for text-to-image tasks. These works have been however so far applied mostly to natural images rather than specialized domains such as solar data, and rarely in setups where the quality of fine-scale details is critical.


\textbf{Applications in solar physics.}
Due to the large data set sizes in solar physics, numerous applications of machine learning and artificial neural network methods have been employed in recent years. Examples are solar flare prediction \cite{Bobra2015}, coronal hole detection \cite{reiss2015}, coronal mass ejection (CME) prediction \cite{Bobra2016}, solar energetic particle event prediction \cite{Kasapis2022}, 3D reconstruction of coronal loops of the solar magnetic field \cite{chifu20213d}, and sunquake detection \cite{Mercea2023}. The interactive HelioML book \cite{HelioML2020} gives a more extended overview of machine learning work in heliophysics.

Since especially solar image data from the SDO mission requires extensive preprocessing to have a uniform dataset available for machine learning, efforts have started to make such datasets available to the broader community, see, e.g., \cite{Galvez2019}, where a large, unified and multispectral dataset of the SDO mission was compiled. However, the resolution of the images in this dataset is limited, therefore, we follow our own data preparation routine, as described in the following section, to enable learning with high resolution. 

Studies that directly use high resolution solar images or even attempt unconditional generative modelling are still rare. Efforts focus on the conditional generation of e.g. AIA coronal data from HMI magnetogram data \cite{Park2019} or magnetograms for the solar far side where no corresponding instruments are available \cite{Kim2019,Sun2022,Jeong2022}. There are also attempts to desaturate overexposed images \cite{Yu2021}, reconstructions of the solar radiation field \cite{Bintsi2022}, and approaches to directly predict flaring activity from full solar disk images \cite{Jonas2018}. However, unconditional noise-to-image synthesis appears to be still understudied in the field of solar physics. We think such a fundamental approach can serve as a testbed for model performance, where lessons learned can be transferred to conditional setups. Furthermore, generative models can help with data augmentation in situations where data availability (e.g. for certain wavelength ranges) is still limited.

\textbf{Image Quality Metrics.}
In practice, the \emph{Fréchet Inception Distance}~\cite{heusel2017gans} is the most frequently applied image quality metric. It employs the features of the  \textit{pool3} layer (dimension 2048) of the pre-trained Inception-v3 neural network. These features are calculated both for real images and generated images and the Fréchet distance between both distributions is calculated assuming each feature distribution is a multivariate Gaussian. Periodic calculation of the FID for monitoring the training progress is implemented and has been useful for this study as well.
However, FID has limitations, (a) it is using the Inception-v3~\cite{szegedy2016rethinking} network which was pre-trained on natural images (ImageNet),  where the data distribution is largely dissimilar to solar data. In addition, the pre-trained feature extractors used in ProjectedGAN are also trained on ImageNet which might give a superficial advantage to ProjectedGAN models' (b) it is well known that FID is biased and the bias can depend on the sample size but also the generative model~\cite{binkowski2018demystifying,chong2020effectively}, which can make comparison between models hard (c) standard FID setup resize images to 299x299, thus it is not ideally suitable to measure quality of fine-scale details.

To overcome (a), we control FID reported performance using a random Inception-v3 network (rFID) and RN50-based CLIP pre-trained network, following~\cite{sauer2021projected}. To overcome (b), we consider the unbiased KID metric~\cite{binkowski2018demystifying} as an alternative to FID. To overcome (c), we consider a patch-based version of FID, where FID is computed on patches at different scales. 

\begin{figure}
\centering
\includegraphics[width=1.0\textwidth]{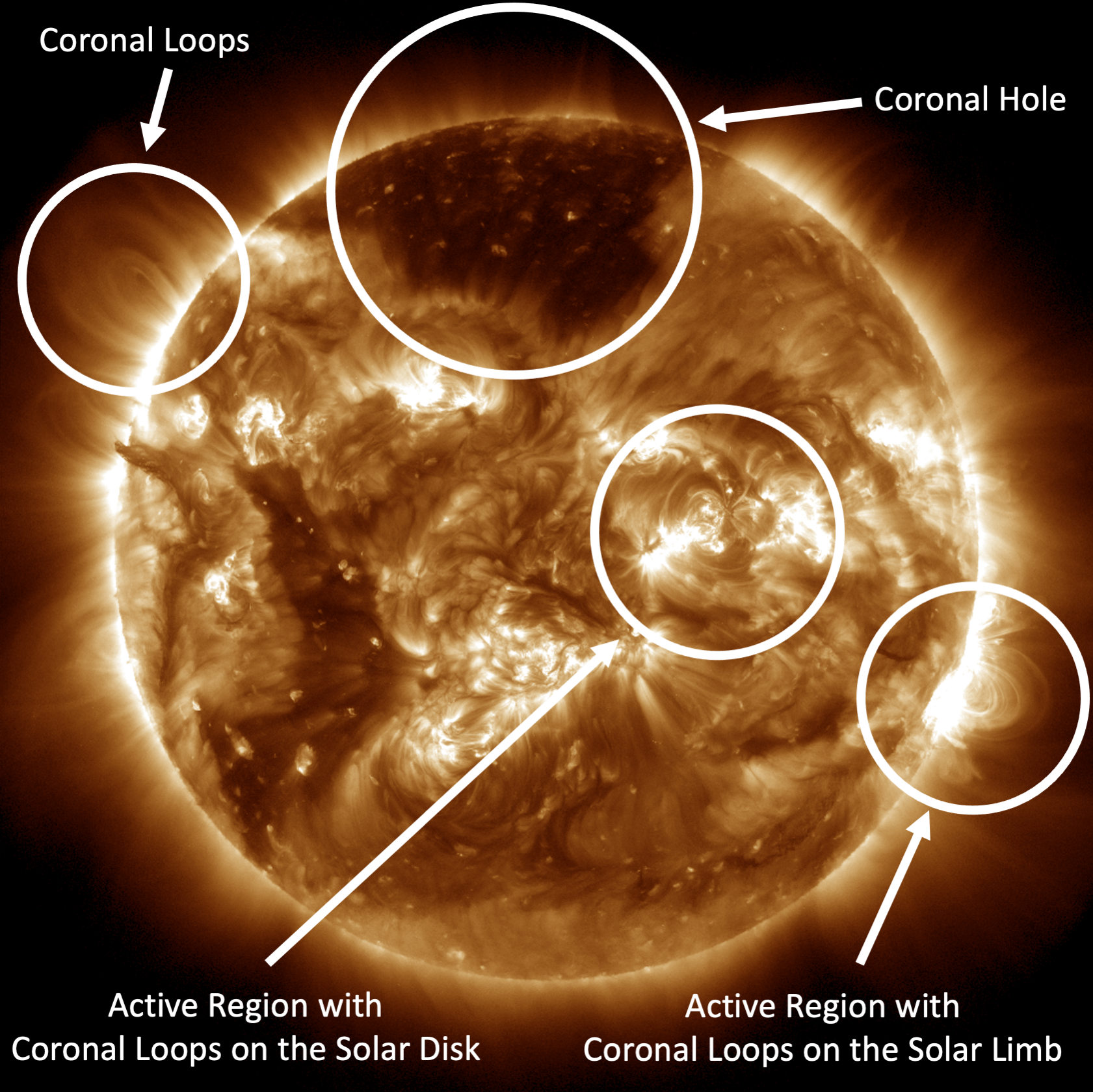}
\caption{Sample of a real EUV solar image in 193\AA~with indications to the most typical prominent features commonly observed, such as coronal holes, active regions and closed coronal loops with many fine scale structures.}
\label{fig:sun_feature}
\end{figure}

Next to quantitative image quality metrics such as FID, we found that visual inspection can guide the investigation as well. Key features are also accessible to the non-experts (a) the roundness of the solar disc and (b) closed loop-like structures in the solar atmosphere, so called coronal loops. See Fig.~\ref{fig:sun_feature} for a visual account of these features. While round discs could be obtained in many cases, especially the existence of coronal loops in generated data turned out to be the key indicator of good fine scale image quality. The presence or absence of these structures in generated data that can be found in most images with high solar activity is the qualitative characteristics that we found to leave the smallest ambiguity.

In Sec.~\ref{sec:experts}, we report on feedback from human subjects not connected to our study on the results we have perceived as being best. The reader is kindly referred to the supplemental material in Appendix~\ref{appendix_images} for more detailed impressions of the generated images. 




\section{Dataset}
We base our study on a subset of Extreme UltraViolet (EUV) images from the SDO AIA instrument \cite{Lemen2012SoPh}. SDO takes data from Earth orbit since February 2010, offering an unprecedented, very large dataset of solar images in different optical and EUV spectral bands, capturing solar atmospheric structures in high resolution and with excellent coverage and temporal cadence of one image every 12 seconds. It provides high resolution solar observations (up to 4096x4096 pixels). This accounts for about 1TB of data per day for almost one full solar cycle. The mission also carries a Helioseismic and Magnetic Imager (HMI) instrument that produces spatially resolved doppler and magnetogram images with the same spatial resolution and similar temporal cadence as AIA. Each EUV spectral range explores different heights in the solar chromosphere and corona, hence is capable of detecting different structures of interest that contain complementary information about the state of the solar atmosphere.

Data from SDO is available in different formats and at different processing levels. We base our dataset on Level 1 data that can be obtained, for instance, through the SunPy\footnote{\url{https://sunpy.org/}} Federated Internet Data Obtainer (FIDO) API. The data is downloaded in the Flexible Image Transport System (FITS) format standard, containing the image pixel intensities together with a large set of metadata.  

We process the image data with routines provided by the AIA team in the python package AIApy\footnote{\url{https://aiapy.readthedocs.io/}} \cite{Barnes2020}, which includes scaling the images to a common plate scale and correcting for instrument degradation effects. Our pipeline is similar to the procedure described in \cite{Galvez2019}.

For our generative training, we use a dataset of 40K images spread evenly across the SDO mission duration to cover all solar activity levels. We filter non-valid images\footnote{The raw FITS images have a metadata "Quality" field, we only keep valid ones with Quality=0, see \url{https://stanford.io/3R78mVV}}. The raw data has pixel intensities ranging from 0 to 16383 where most of the mass is concentrated between 0 and ~500; there are as well negative pixel values which are considered as instrument measurement errors. We preprocess the data by clipping the pixel values to 1 as a minimum value, then applying a (natural) log transform followed by normalization to $[0,1]$ by dividing by the maximum value. We train our models on AIA 193\AA\ subsampled images of size 1024x1024. For visual inspection, real and generated images are colorized with the colormap \texttt{sdoaia193} from the SunPy package  to align with the most common visualization of the data. 

\section{GAN models experiments}
\label{sec:GAN}

\subsection{Setup and training procedure}
\label{subsec:GAN:setup}


Here we describe setup of series of experiments conducted using different GAN models in our attempt to generate high quality solar image samples, following previous works that report high quality generation for various natural images. 

\begin{figure*}[t]
\begin{center}
\includegraphics[width=1.0\textwidth]{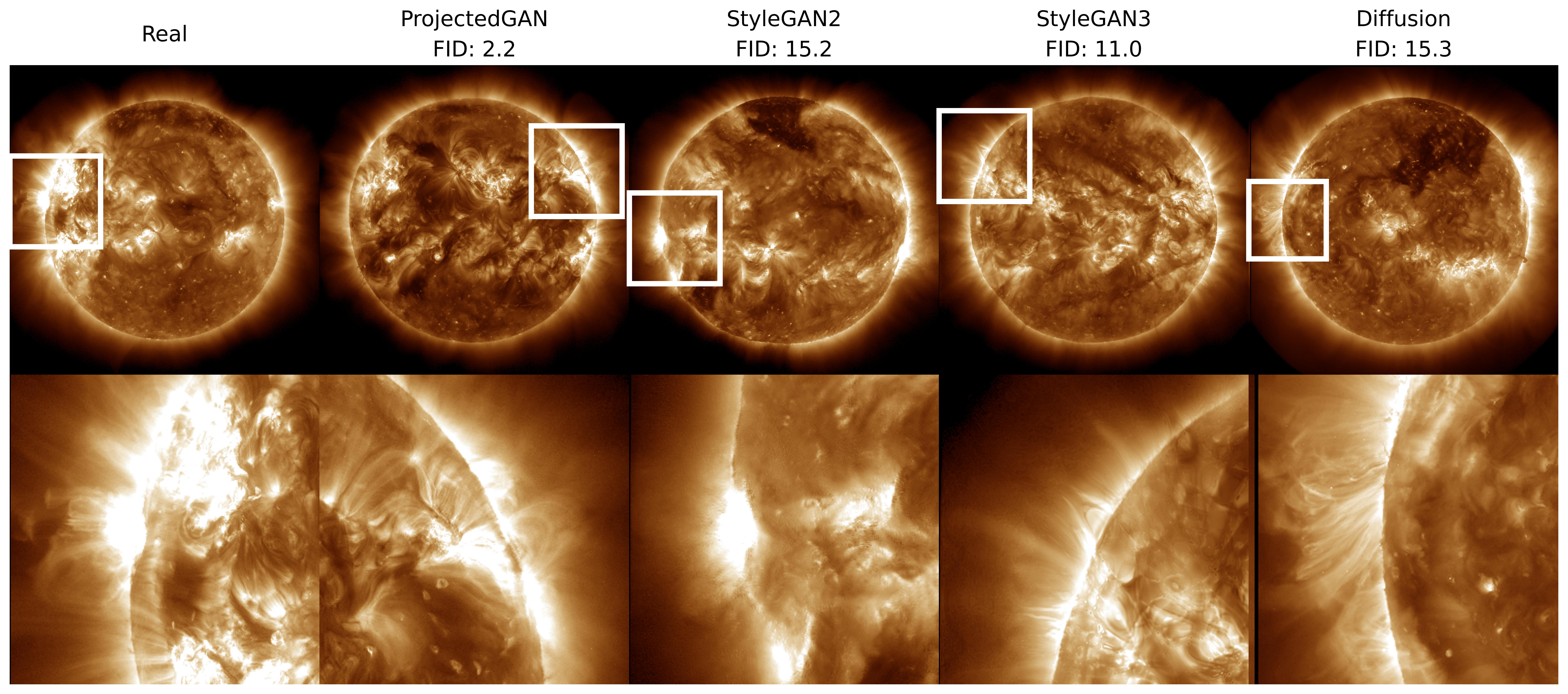}
\caption{
  Image quality comparison between (from left to right) a real image, an image generated with ProjectedGAN, with StyleGAN2 and StyleGAN3, and a diffusion model (ADM). Below each image, a zoomed in version of the region indicated with the white box is displayed. Only ProjectedGAN and ADM can reproduce coronal loops, the reentrant structure in the solar atmosphere. StyleGAN2 and StyleGAN3 cannot reproduce these features. 
}
\label{fig:gan_comparison}
\end{center}
\end{figure*}

In a series of experiments, we applied the published implementation of StyleGAN2-ADA, StyleGAN3, ProjectedGAN and StyleGAN-XL to our prepared dataset. All implementations are based on the StyleGAN2-ADA implementation by NVIDIA\footnote{\url{https://github.com/NVlabs/stylegan2-ada-pytorch}}. We adjusted the respective implementations for the usage on the supercomputer, and faithfully reproduced the computational setup following previous work. This involves introducing a multi-node launching procedure, as an 8-GPU-setup as chosen in the previous work requires the usage of two compute nodes. The implementation periodically performs an FID evaluation of generated images. In agreement with visual inspection, we report the generated images with the lowest FID as the best images in every experiment. 


Motivated by the failures of the initial experiments, we suspected the relative dynamics of generator and discriminator learning process to be the key reason. Therefore, we investigated different hyperparameter setups with separately varied learning rates and number of optimization steps for generator and discriminator for ProjectedGAN. Especially, a reduced generator learning rate would, in many cases improve the result. However, we also observed a strong fluctuation in the FID measured after training. Due to the computational effort, we repeated only our best ProjectedGAN configuration (Suppl. Tab.~\ref{table:hyperparam}) five times (Suppl. Fig.~\ref{fig:fid_variation}). 

\subsection{Results}
\label{subsec:GAN:results}

Our main results obtained with ProjectedGAN, StyleGAN2-ADA and StyleGAN3 are shown in Fig.~\ref{fig:gan_comparison} along with real images. The mean FID measured in five ProjectedGAN runs is evaluated to 4.2 with a standard deviation of $\pm 2.0$. In the panel, we show a generated sample for a run with the best obtained FID of 2.2. Visually, the quality of the different runs is hard to distinguish. StyleGAN2-ADA and StyleGAN3 create images of which the FID is evaluated to 15.0 and 11.0 respectively. While ProjectedGAN's suns are round, StyleGAN2's suns deviate from circular shape. StyleGAN3's suns are better, yet deformation is visible. The quality of fine scale details is very different. For StyleGAN2 the visual image quality appears coarse, while StyleGAN3 shows intricate details but with very visible artifacts. Both models do not produce visual structures resembling coronal loops. ProjectedGAN's visual features are very fine and clearly also show coronal loops, and even for the expert, it is difficult to separate real from fake images. None of the training processes shows signs of mode collapse.

\begin{figure*}[h]
  \includegraphics[width=\textwidth]{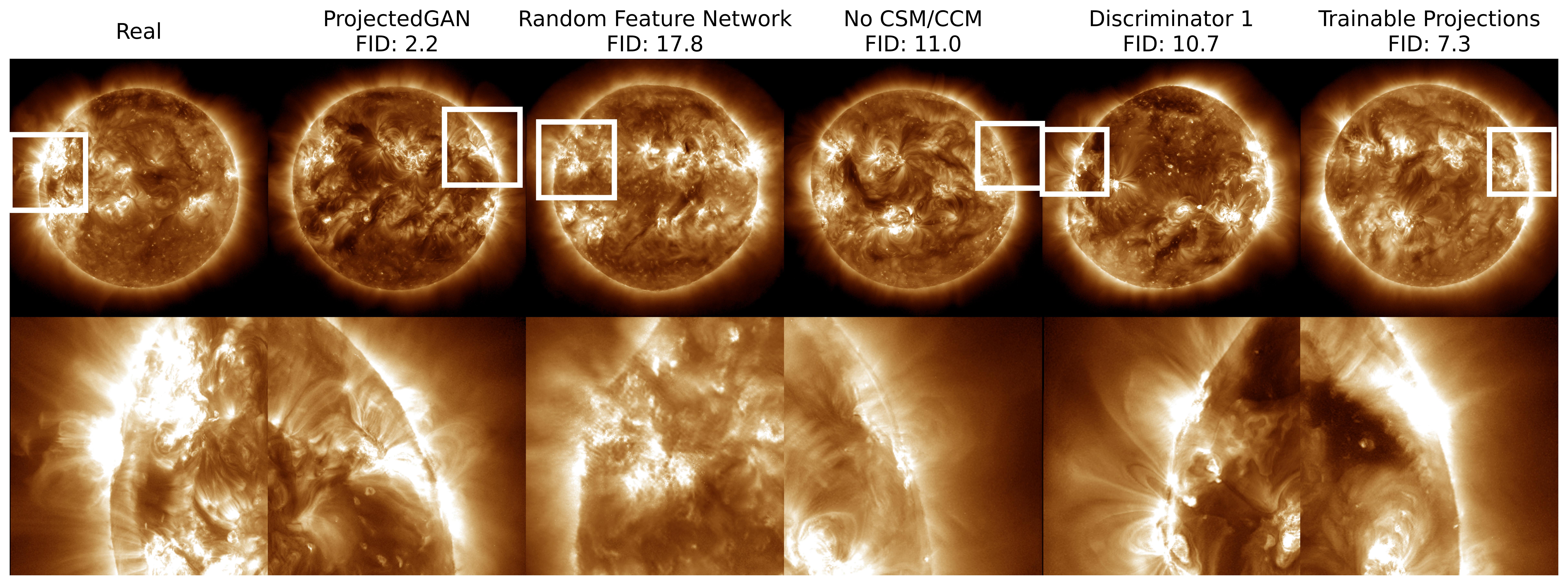}
  \caption{
  Qualitative Overview of the ProjectedGAN ablation studies. The original images are displayed on the left and different ablation studies beneath. Each column displays representative example of (top) the solar disc and (bottom) a coronal loop region.
  }
\label{fig:projectedgan}
\end{figure*}

In order to understand this behaviour better, we have performed a series of ablations which systematically investigate which innovations of ProjectedGAN contribute to this breakthrough. This ablation study is summarized in  Table \ref{table:projectedgan} and selected results are shown in Fig. \ref{fig:projectedgan}. ProjectedGAN comes with three main innovations that all affect the structure of the discriminator. On the one hand, a frozen pre-trained network is used to extract features that are subsequently passed on. To assess the necessity of pre-training we replaced the pre-trained weights with randomly initialized weights. To our surprise, this led the solar shape intact and generated reasonable coarse features, but led to significant quality reduction on the fine scale, consistent with an FID of 17.8. Furthermore, we investigated if a change in the feature network depth affects the image quality. While the ProjectedGAN authors observe a slight deterioration of image quality, in our case replacing the EfficientNet-Lite0 by respectively deeper variants Lite1, Lite2 and Lite3 does not affect the image quality (see Suppl. Fig. \ref{table:metrics}).


\begin{table}[h]
\small
\begin{center}

\begin{tabular}{lrcc}
\toprule
                                            Run &           FID $\downarrow$ & SR & CL \\
\midrule
ProjectedGAN (Baseline) & 4.2 $\pm$ 2.0 &  $++$ & $++$  \\
 Random feature network &          17.8 & $++$ & $0$\\
                             EfficientNet-Lite1 &           7.4 & $++$ & $++$\\
                             EfficientNet-Lite2 &           4.1 & $++$ & $++$\\
                             EfficientNet-Lite3 &           3.8 & $++$ & $++$\\
                          No Cross Scale Mixing &           9.6 & $+$ & $+$  \\
     No Cross Scale/Channel Mixing &          11.0  & $+$ & $+$ \\

                    Discriminator 1 &          10.7 & $0$ & $+$ \\
                              Discriminator 1,2 &           7.4 & $+$ & $+$ \\
                            Discriminator 1,2,3 &           6.5 &  $+$ & $++$ \\
                              Augmentations off &           4.2 & $++$ & $++$\\
                          Trainable Projections &           7.2 &$++$ & $++$\\
           
\bottomrule
\end{tabular}
\end{center}
\caption{ProjectedGAN ablations. The columns SR and and CL indicate results from qualitative visual inspection regarding sun roundness (SR) and presence of coronal loops (CL). A zero   indicates unsatisfactory performance, two plus symbols indicate that deviations from roundness is very difficult to see or that coronal loops are clearly visible. A single plus sign indicates that artefacts a still clearly visible.}
\label{table:projectedgan}
\end{table}

Furthermore, ProjectedGAN employs a pair of techniques called Cross-Channel and Cross-Scale Mixing (CCM and CSM). CCM embeds the feature channels of each convolutional output into a higher-dimensional space. CSM mixes these feature representations from coarser into finer scales by interpolating and summation. We observe that especially CSM leads to a considerable improvement of image quality. If either projection method is omitted, the solar roundness is not as good, and the quality of fine-scale details deteriorates. Coronal loops remain present but exhibit clear artifacts. 

Additionally, ProjectedGAN employs multiple discriminators, i.e.~the mixed features are digested by multiple networks with independent discriminator losses of equal functional form. This structure of multiple discriminators ensures that features on all scales can be employed independently. We investigate how using only one, only two and only three discriminators affect the result. Our results clearly indicate an improvement of the image quality by each discriminator. Interestingly, by using only D1, the discriminator that processes only fine-scale features directly, and coarse features only through multiple levels of cross-scale mixing, produces suns with appealing fine-scale scale features but clear deviations from spherical shape. 

Ultimately, we investigated if disabling the applied augmentation has influence on the obtained results and if training the parameters of the projections influences the result. The first can be ruled out, and in the second case we observe a slight decrease in image quality, with FID dropping to 7.3.

\section{Diffusion models experiments}
\label{sec:difusion}
\subsection{Setup and training procedure}
\label{subsec:diffusion:setup}

 We follow ADM~\cite{dhariwal2021diffusion}, and use their UNet architecture based on convolutional residual blocks and global attention at different resolutions and adapt OpenAI's implementation of ADMs\footnote{\url{https://github.com/openai/guided-diffusion}}. We experiment with different denoising steps (250, 500, 1000, 2000, 4000) and train the models for a maximum of 100K training steps with a learning rate of $0.0001$ and a batch size of $64$. For sample generation, we use 250 steps to make generation faster.

\subsection{Results}
\label{subsec:diffusion:results}

For diffusion models, the best FID we obtain is $15.3$ (see Fig.\ \ref{fig:gan_comparison}) with 1000 training denoising steps and 250 sampling steps.
Interestingly, even if the FID is much higher than the best ProjectedGAN results ($\text{FID}=2.2$) obtained in Sec.~\ref{subsec:GAN:results}, we observe that the best diffusion model is able to generate a correct spherical shape, and high quality fine-scale details, e.g., we observe that it can successfully reproduce coronal loops (see also Suppl. Fig. \ref{fig:comparison}).
To make sure ProjectedGANs do not have an advantage in FID due to the fact that it is using ImageNet pre-trained features on the discriminator, we follow ~\cite{sauer2021projected} and also compute rFID and CLIP-RN50. 
We find that the order of different models (ProjectedGAN ablations in Sec.~\ref{subsec:GAN:results} together with diffusion models) is consistent. We find a Spearman rank correlation of $\rho=0.75$ 
 between FID and rFID, and $\rho=0.94$ between FID and CLIP-RN50, also ProjectedGANs have consistently smaller FID compared to diffusion models (see also Suppl. Tab. \ref{table:metrics}, \ref{table:metrics-patch}, \ref{table:metrics-corr}).

\section{Assessing quality by human experts study}
\label{sec:experts}
We conducted a small exploratory study with a sample of 20 human subjects with different proficiency in solar physics research. On a scale from 1 to 5 they rated their expertise with an average of 2.6 points and a standard deviation of 2.42. We showed them a sample of 5 real and 5 fake samples generated from both the best GAN results and the diffusion models. The aim was to test if humans can distinguish real from fake samples significantly better than random guessing.

\begin{figure*}[h]
\begin{center}
  \includegraphics[width=0.79\textwidth]{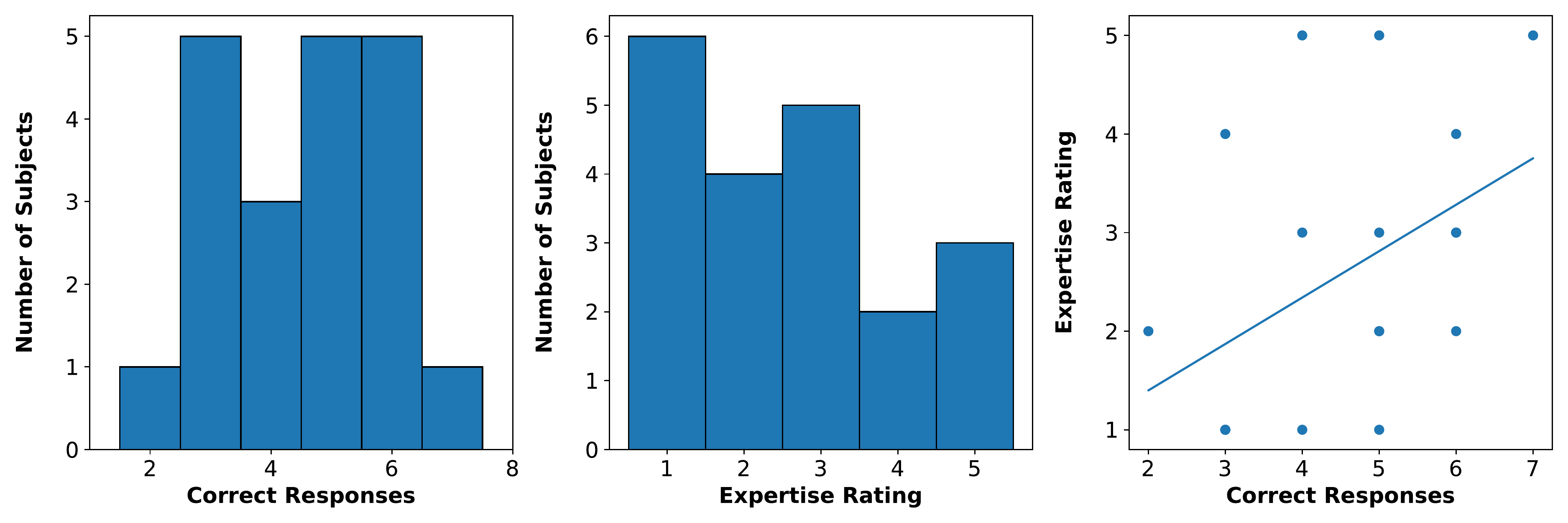}
  \caption{
  Histograms of the number of correct responses (out of 10 questions) from the human expert study (left) and their expertise self assessment (on a scale 1-5, middle). The correlation between both is shown in the right panel with a correlation coefficient of 0.46.
  }
  \label{fig:human}
  \end{center}
\end{figure*}

The average score of the entire group of 20 subjects was 4.55 with a standard deviation of 1.39. The histogram of the number of correct responses is shown in Fig.~\ref{fig:human}. The result is consistent with the hypothesis of random guessing, with an aggregate two-sided p-value of 0.66. The data indicates a small correlation between the expertise rating and the number of correct responses. Overall, our sample size is too small to allow for firm conclusions but we view this as indication that humans, even with expertise in solar EUV images, struggle to identify fake images reliably. 

\section{Discussion}
\textbf{Generation of fine-scale details.} 
Going through intensive experiments with various GAN approaches, we finally obtain with ProjectedGAN solar images that have comparable good quality to images produced in the experiments with the diffusion model. We clearly see that no single technique introduced by ProjectedGAN can be made solely responsible for the observed improvement over less successful GAN models with basic StyleGAN architecture. To achieve good quality, following components turn out to be necessary from the conducted ablation study: the pre-trained feature network, feature mixing across scales and independent discriminators for all scales. This clearly underlines the importance of the employed discriminator architecture. 

We observe that architectures like StyleGANv2 and StyleGANv3, that do not possess such explicit mechanisms to deal with multi-scale nature of the image signal built into discriminator, fail to generate necessary fine-scale structures in the solar images. Remarkably, natural images like faces do not pose such a difficulty for the models with basic StyleGAN architectures that struggle on solar image data (Suppl. Fig.~\ref{fig:faces}). This hints on certain properties of the dataset or single image signal statistics that make fine-scale detail learning harder for GAN models without discriminator mechanisms for explicit multi-scale treatment. Moreover, as methods known to stabilize training and avoid mode collapse by introducing differentiable augmentation like DiffAug or StyleGAN-Ada alone did not manage to fix the fine-scale generation in basic StyleGANs, we assume that training instabilities and mode collapse are not the main driving force behind fine-scale corruption here. In addition to multi-scale mechanisms, the importance of having a fixed, pre-trained feature extractor is evident, as either starting with fixed random weights or unfreezing the feature extractor destroys good fine-scale generation despite multi-scale mechanisms still in place.

Contrary to the efforts necessary to get fine-scale generation working well in GANs, standard diffusion model (ADM) operating in pixel space and employing U-Net works without extensive tuning. This is in line with the already observed benefits of diffusion models over GANs, and hints on advantage of multi-step denoising generation methods for proper fine-scale handling, as opposed to single-step one pass generation employed in GANs.

\textbf{Comparing different generative models.} 
When comparing the quality of solar image samples generated by different models, we notice various degrees of degradation either on fine- or coarse-scale level. For instance, for fine-scales, we see particular salient solar image features, like coronal loops outside, or in the active regions within solar disk or on the solar limb, either clearly expressed, or corrupted or entirely gone, depending on model quality. On the coarse-scale level, easily detectable is the preservation of the ideally spherical shape of the solar disk or its distortion.

As we measure FID, the scores obtained for the GAN models seem on the one hand to be well ordered according to the observed quality of fine and coarse scales of the solar images in correspondence with the trained model quality. The lowest FID scores observed correspond clear coronal loop structures at the solar limb and on the solar disk at fine scale as well as to the spherical shape of the sun disk at coarse scale. ProjectedGANs achieve the lowest FIDs. Ablations that corrupt either fine or coarse scale result in higher FIDs, where we see distortion or loss of coronal loop structures and distortion of the solar disk shape. Poor models with highest FID scores like StyleGANv2 family do not posses any proper fine scale structure at all, keeping though a proper coarse round shape of the solar disk.

On the other hand, we observe that despite being the best among assessed image quality on fine and coarse scale, diffusion models obtain higher FID than generated samples obtained by GANs which have poorer quality upon visual inspection. This inconsistency cannot be explained solely by the fact that we use a feature extractor pre-trained on ImageNet in ProjectedGANs, which would give an advantage in terms of FID, since Inceptionv3 is also trained on ImageNet. We find consistently better performance for ProjectedGANs when using other metrics (rFID, CLIP-RN50) and similar model ranking in general (Suppl. Tab. \ref{table:metrics}, \ref{table:metrics-patch}, \ref{table:metrics-corr}). After checking the pixel distribution, we observe that the  histogram of pixel intensities of diffusion (ADM) samples do not match the histogram of real data (see Suppl. Fig.~\ref{fig:histogram}): the real data has heavier tails at the left side (small pixel values) and diffusion samples have heavier tails on the right side (large pixel values), while the one obtained with ProjectedGAN matches the real data much better. This distribution mismatch may be the reason for higher FID scores and the inconsistency observed. Again, this calls for caution when judging generated sample image quality - FID alone cannot give a comprehensive answer, and other domain-specific scores or visual inspection by experts might be necessary, as it is this case in our study for solar images.


%
%

When conducting the human experts study, we were then taking those sample images that showed high quality on both fine and coarse scale -  generated by ProjectedGAN and ADM. Our study with human experts confirms the quality of generated samples - as human were not able to distinguish reliably real from generated solar images.





\textbf{Limitations of this study} While we took various GAN models for the comparative study and covered a broad range of possible architectures, we did not have enough computational resources to conduct a dense systematic hyperparameter search for tuning the training procedure. We therefore cannot exclude the possibility that further extensive tuning would enable basic StyleGAN models to get fine-scale details of sufficient quality on solar images. However, it is clear that models like ProjectedGAN or ADM did not require such tuning to work well, providing evidence for the benefits of explicit multi-scale mechanisms in GANs and for multi-step diffusion generation. Further,  diffusion based training as used in ADM leads to slow inference compared to GANs. Approaches like latent diffusion \cite{rombach2022high} training in latent instead of pixel space substantially decrease inference time.  Pre-training on large-scale generic image datasets may further have increased the capability of feature extractors like used in ProjectedGAN. Finally, we train here only on static solar images of a specific wavelength without taking into account temporal or multi-spectral information that does exist in the SDO dataset.

\section{Conclusion and Outlook}
Encouraged by the previous works showing capability to generate high quality high resolution natural images using architectures like StyleGAN, we started our study with initial expectation to generate similar high quality high resolution samples from the SDO dataset containing solar images using same methods. However, we observe that basic StyleGAN architectures and their extensions like StyleGANv2, StyleGAN-ADA, DiffAug and StyleGANv3 are not capable of generating solar images of sufficient quality, failing at crucial fine-scale details - which is not observed in this way for the scenario of natural image generation. This calls for caution when applying generative approaches highly successful on natural image data to scientific image signals. By executing extensive experiments, we find that ProjectedGAN with its pre-trained feature extractor, cross scale mixing and multi-scale discriminators provides solar image samples with high quality on both fine and coarse scale. Diffusion-based ADM can also achieve comparable high sample quality without tuning effort. Samples created by both methods are found to be indistinguishable from real data in a human expert evaluation experiment.

As observed in this study, the scientific image dataset scenario may differ from standard requirements for natural image generation. To accelerate further progress in direction of generative modeling for high resolution scientific data, we open-source the outcomes of this work. For solar image modelling, exploring latent space of trained models, and using further information like temporal, multi-spectral and textual meta data available from SDO are future directions leading to powerful, physics-aware generative models for solar state interpretation and prediction.

\section{Acknowledgements}

We would like to thank Ruggero Vasile for his contributions during the initial phases of the project with StyleGAN2 models, including data gathering, data pre-processing and preparation, as well as Yuri Shprits for support in the begining of the project. We would like to also thank Katja Schwarz for all her insights and experiment suggestions on our ProjectedGAN ablations.

This work is supported by the Helmholtz Association Initiative and Networking Fund under the Helmholtz AI platform grant and the HAICORE@JSC partition.  The authors gratefully acknowledge the Gauss Centre for Supercomputing e.V. (www.gauss-centre.eu) for funding this project by providing computing time through the John von Neumann Institute for Computing (NIC) on the GCS Supercomputer JUWELS at the J\"ulich Supercomputing Centre (JSC). Partial support of DFG (SFB 1491) is acknowledged.


\bibliography{main}
\bibliographystyle{ieee_fullname}
\newpage
\input{appendix}

\end{document}

%% file: appendix.tex
\onecolumn
\begin{appendix}

\begin{center}
{\Large\bf Supplementary}
\end{center}

\section{Additional details on solar image sample quality comparison}
\label{appendix_images}
\begin{figure}[h]
  \includegraphics[width=\textwidth]{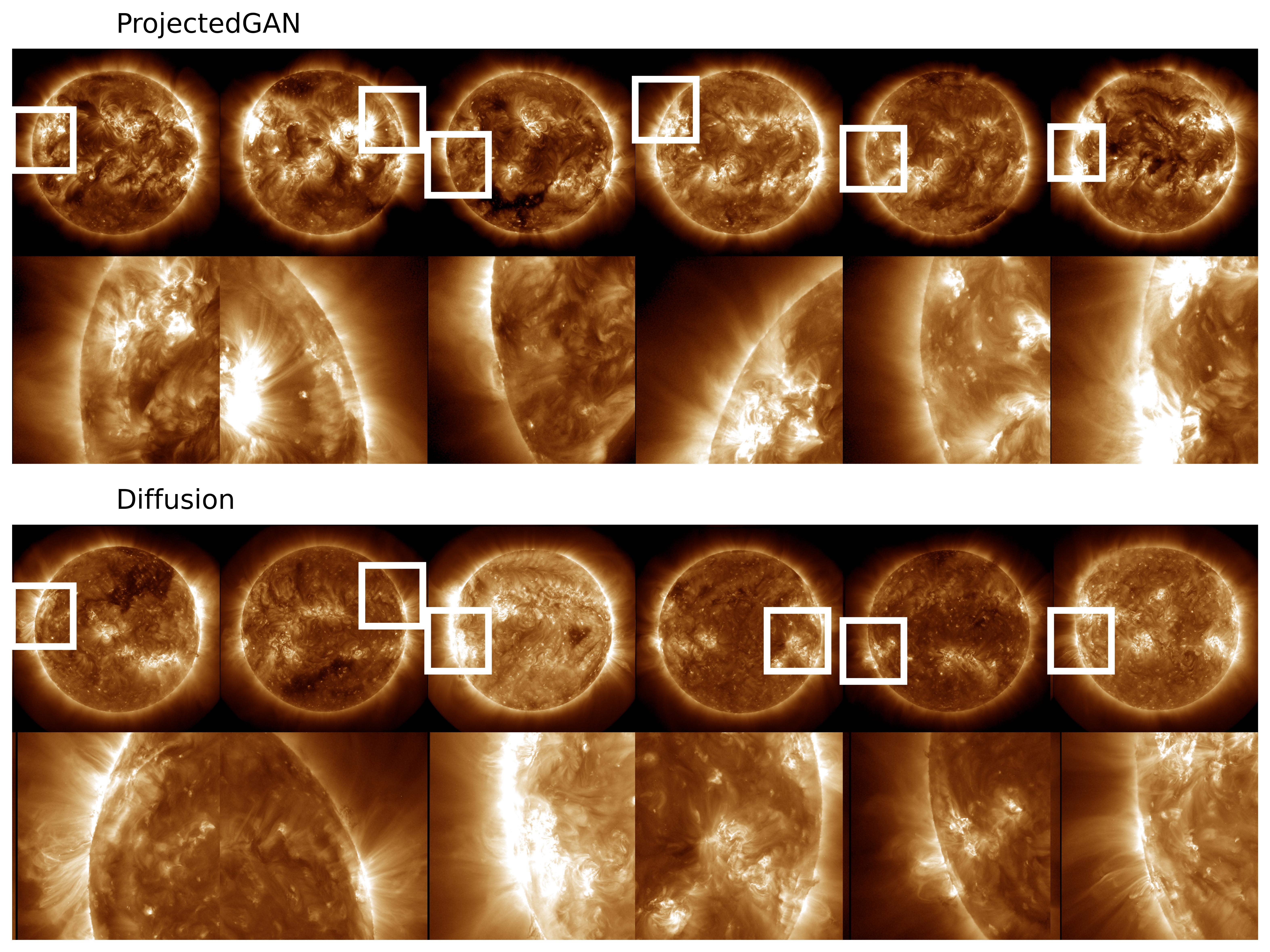}
  \caption{
  Comparison of ProjectedGAN results and Diffusion model results. Despite the FID being much lower for the ProjectedGAN model (FID 2.2) than the Diffusion model (FID 15.2), both coarse and fine scale agree when evaluated by a human. Both models generated sun of which the shape can visually not be distinguished from a circle. Also fine-scale details are of comparable quality. 
  }
\label{fig:comparison}
\end{figure}

Here we provide an overview over a wider spectrum of generated images with the best models of both classes. Fig.~\ref{fig:comparison} shows five images generated by ProjectedGAN and by the diffusion model (ADM) for different solar activity levels. The insets zoom into regions with coronal loop structures. For both models, the subjective impression is very good. The solar shape is indistinguishable from circular and coronal loops are clearly visible. However, the FIDs obtained for both models are significantly different, being 2.2 for ProjectedGAN and 15.2 for the diffusion model, showing the FID alone cannot properly reflect image quality. 

\section{Additional details on GAN experiments}
\subsection{Hyperparameters}
In this section, we report the hyperparameters employed for ProjectedGAN in the configuration with which we obtained optimal results. The reported parameter names are inspired by the parameters of the ProjectedGAN training code\footnote{\url{https://github.com/autonomousvision/projected-gan}}, however for readability, abbreviations have been expanded. 

\begin{table}[h]
\small
\begin{center}
\begin{tabular}{lll}
\toprule
Group & Parameter & Value \\
\midrule
Generator & \texttt{type} & StyleGAN2 \\
 & \texttt{z\_dim} & 64 \\
 & \texttt{w\_dim} & 128 \\
 & \texttt{num\_mapping\_layers} & 2 \\
G-Optimizer & \texttt{type} & Adam \\
 & \texttt{betas} & 0,0.99 \\
 & \texttt{learning\_rate} & 0.0005 \\
D-Optimizer & \texttt{type} & Adam \\
 & \texttt{betas} & 0,0.99 \\
 & \texttt{learning\_rate} & 0.002 \\
Training & \texttt{batch\_size} &  32 \\
 & \texttt{num\_gpus} &  8 \\
Projection & \texttt{diffaug} & True \\
 & \texttt{type} & 2 (CSM+CCM) \\
 & \texttt{out\_channels} & 64 \\
Discriminator & \texttt{num\_discs} & 4 \\
\bottomrule
\end{tabular}
\end{center}
\caption{ProjectedGAN hyperparameters of the baseline run. The evaluation of multiple runs with this hyperparameter set can be found in Fig.~\ref{fig:fid_variation}.
}
\label{table:hyperparam}
\end{table}

\subsection{FID variation}
\begin{figure}[h]
\begin{center}
  \includegraphics[width=0.5\textwidth]{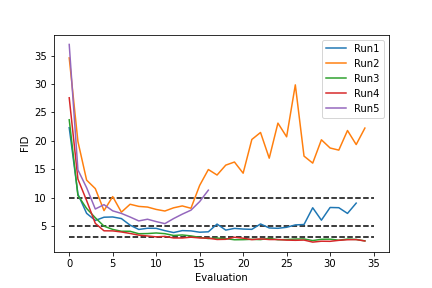}
  \caption{
  FID in the course of training of ProjectedGAN. The panel shows five runs with identical parameters. Horizonal lines at 3,5, and 10 to guide the eye.
  }
  \label{fig:fid_variation}
 \end{center}
\end{figure}

With our studies, we observe that different GAN training runs with identical parameters can lead to very different behaviour even with identical hyperparameters as given in Tab.~\ref{table:hyperparam}. Fig.~\ref{fig:fid_variation} shows the FID in the course of five identical ProjectedGAN runs evaluated every 6400 iterations with a batch size of 32. The five runs exhibit significantly different behaviour. In two runs the FID decreases continuously and subsequently fluctuates between 2 and 3. In three runs, the FID reaches a pronounced minimum and starts increasing again. The resulting subjective image quality of all runs is comparable.

\begin{figure*}[!t]
\begin{center}
\includegraphics[width=0.85\textwidth]{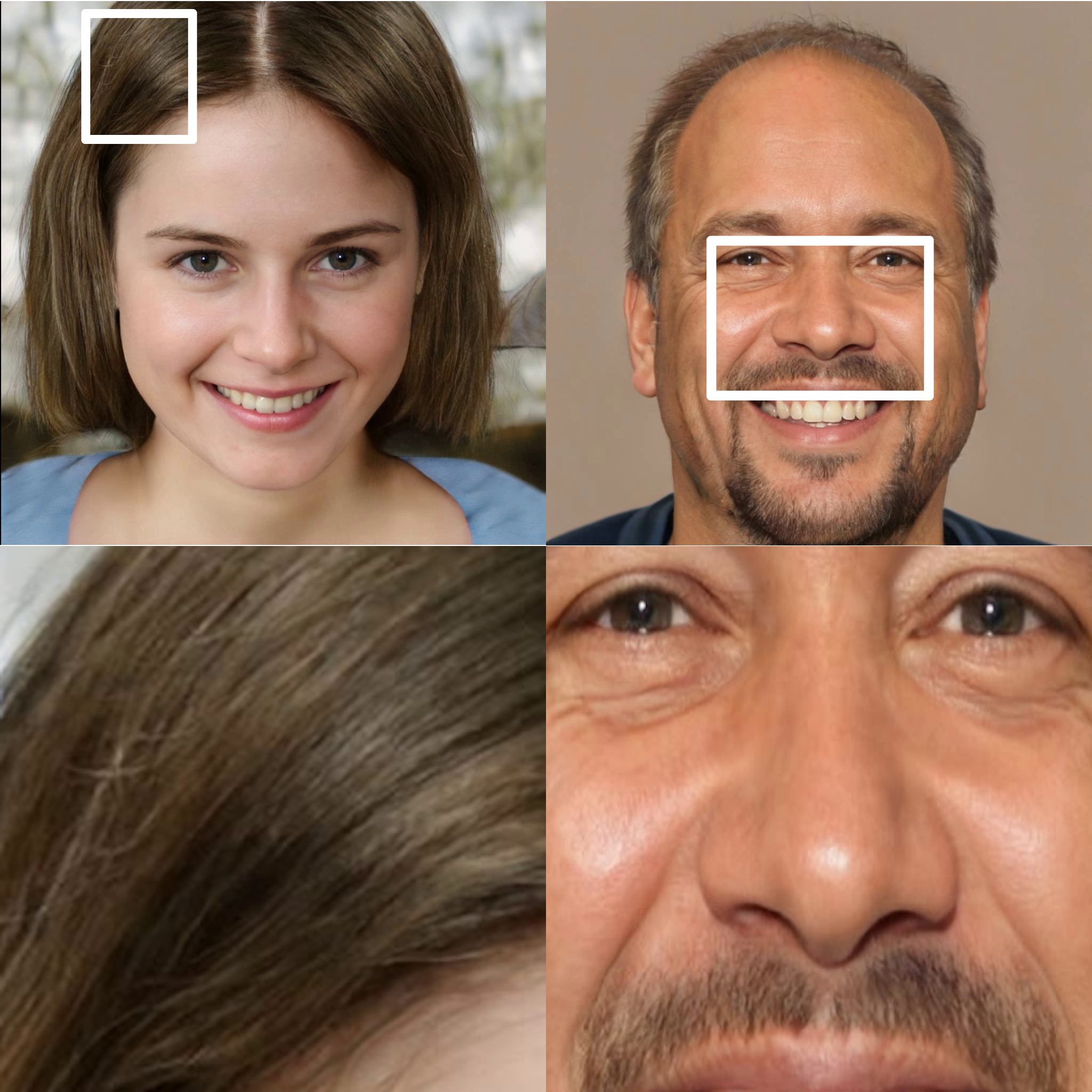}
  \caption{
        In our baseline experiments with FFHQ (human faces) using StyleGANv2 with differentiable data augmentation~\cite{zhao2020differentiable,karras2020training}, contrary our experiments with solar data (see Fig.\ref{fig:gan_comparison}) the model is able to
        generate fine-scale details, e.g., wrinkles, in the head hair, and in facial hair. In solar data, despite optimizing hyper-parameters such as learning rate (for discriminator and generator), trying different augmentations, training on more epochs, we could not find a setup where generation of fine-scale details is acceptable.
  }
  \label{fig:faces}
  \end{center}
\end{figure*}

\subsection{Comparison to natural image generation}
Here we show results of standard StyleGANv2 training on natural face image generation. Contrary to the severe issues StyleGANv2 has when generating fine-scale details for solar images that we observe in this study (Fig.~\ref{fig:projectedgan}), such issues are absent on natural face images, as  demonstrated in Fig. \ref{fig:faces}. Fine-scale details like hair or eye structures are faithfully reproduced. This exemplifies that differences in the solar images and natural face images dataset and single image statistics used for training are crucial for the properties of the resulting models, either being able or not able to handle fine-scale details depending on dataset nature

\subsection{Latent space control}

\begin{figure}[!h]
\centering
\begin{subfigure}{1.0\textwidth}
  \centering
  \includegraphics[width=\linewidth]{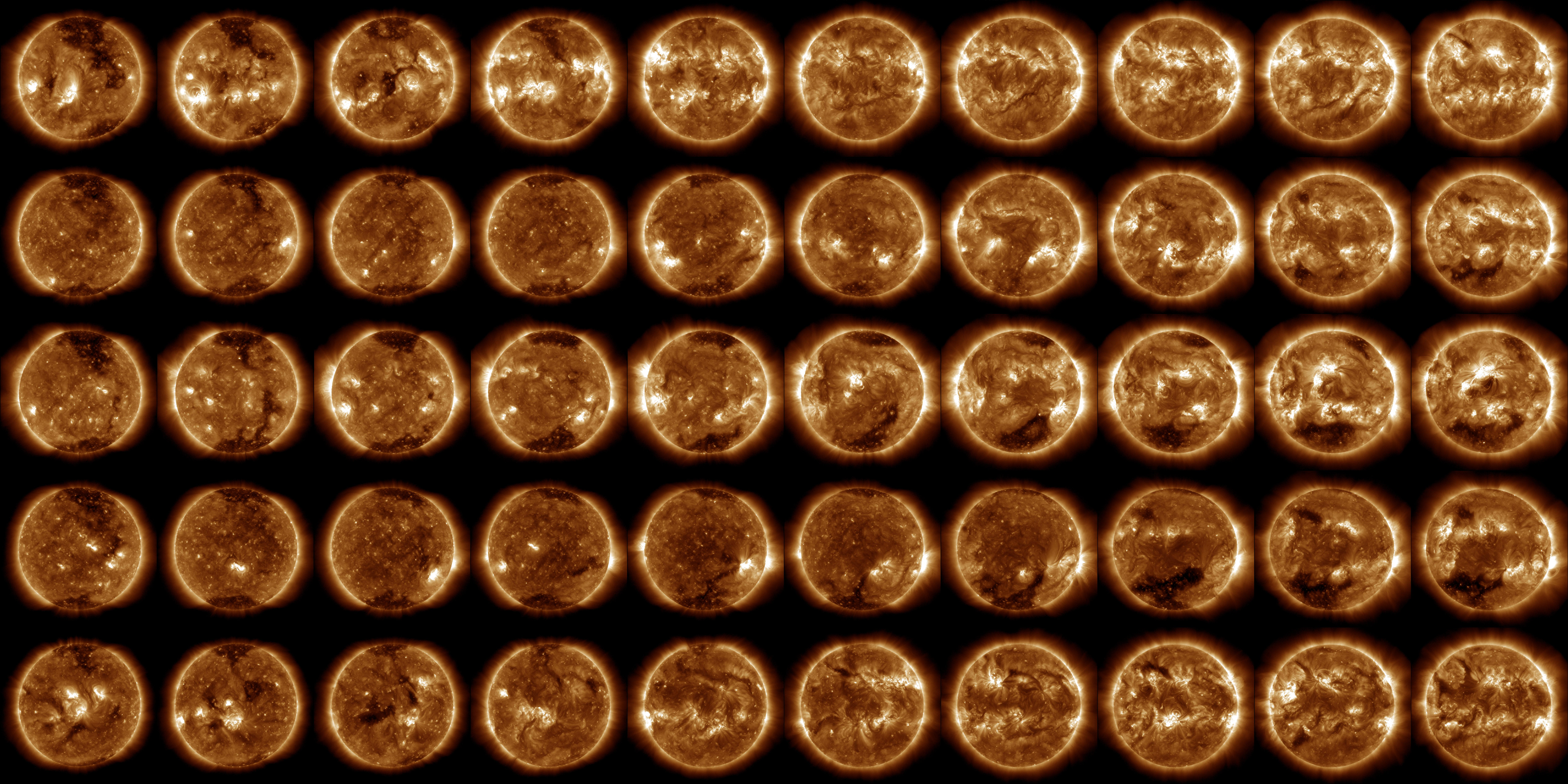}
  \caption{PCA Component 1}
  \label{fig:latent_space_control_1}
\end{subfigure}
\hfill
\begin{subfigure}{1.0\textwidth}
  \centering
  \includegraphics[width=\linewidth]{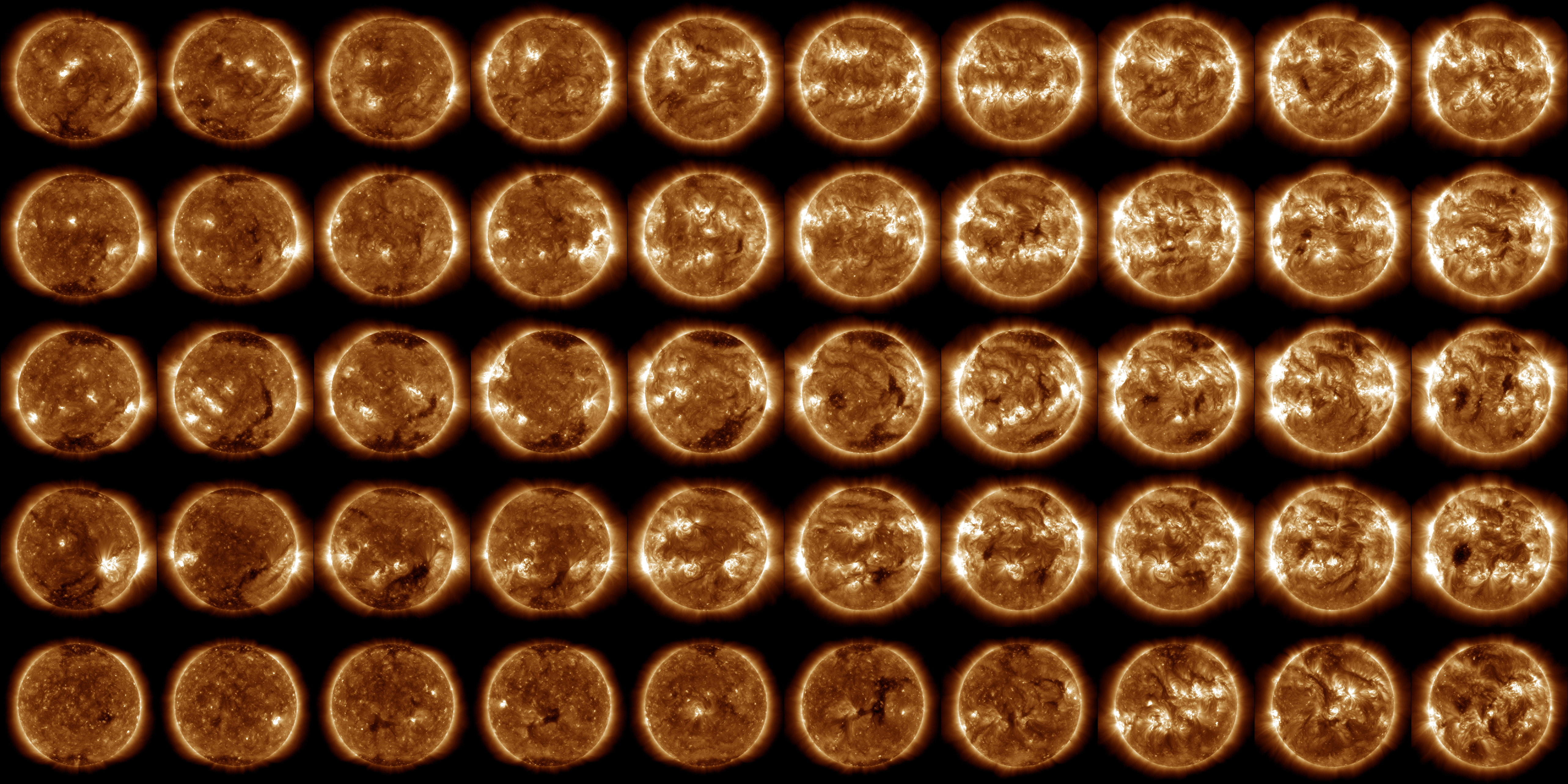}
  \caption{PCA Component 2}
  \label{fig:latent_space_control_2}
\end{subfigure}
\caption{
Unsupervised latent space control based on the \emph{W} space of our best ProjectedGAN model using the GANSpace\cite{harkonen2020ganspace} method. We visualize the the first (a) and second (b) PCA components with highest eigenvalue. Each row is an independent sample and columns  correspond to images obtained by varying of the coordinate of the component.  We observe that the first two principal components span a space where both intensity of sun activity and number of corona holes can be traversed.
}
\label{fig:latent_space_control}
\end{figure}

We conducted an preliminary study on unsupervised extraction of meaningful directions in latent space of the best performing GAN model, ProjectedGAN, following the GANSpace~\cite{harkonen2020ganspace} method. In Fig~\ref{fig:latent_space_control}, we provide a  visualization of the first and second PCA components (based on the \emph{W} space) with highest eigenvalue, where each row is an independent sample, and columns  correspond to variation of the coordinate of the component. It can be observed that the first two principal components span a space where both intensity of sun activity and number of corona holes can be traversed, hinting that latent space already contains a well-shaped useful structure that contains physically meaningful representations of the sun’s state. 

\section{Additional details on diffusion model experiments}
\subsection{Pixel value distribution and FID}
\label{appendix:histogram}
\begin{figure*}[!h]
\begin{center}
\includegraphics[width=0.6\textwidth]{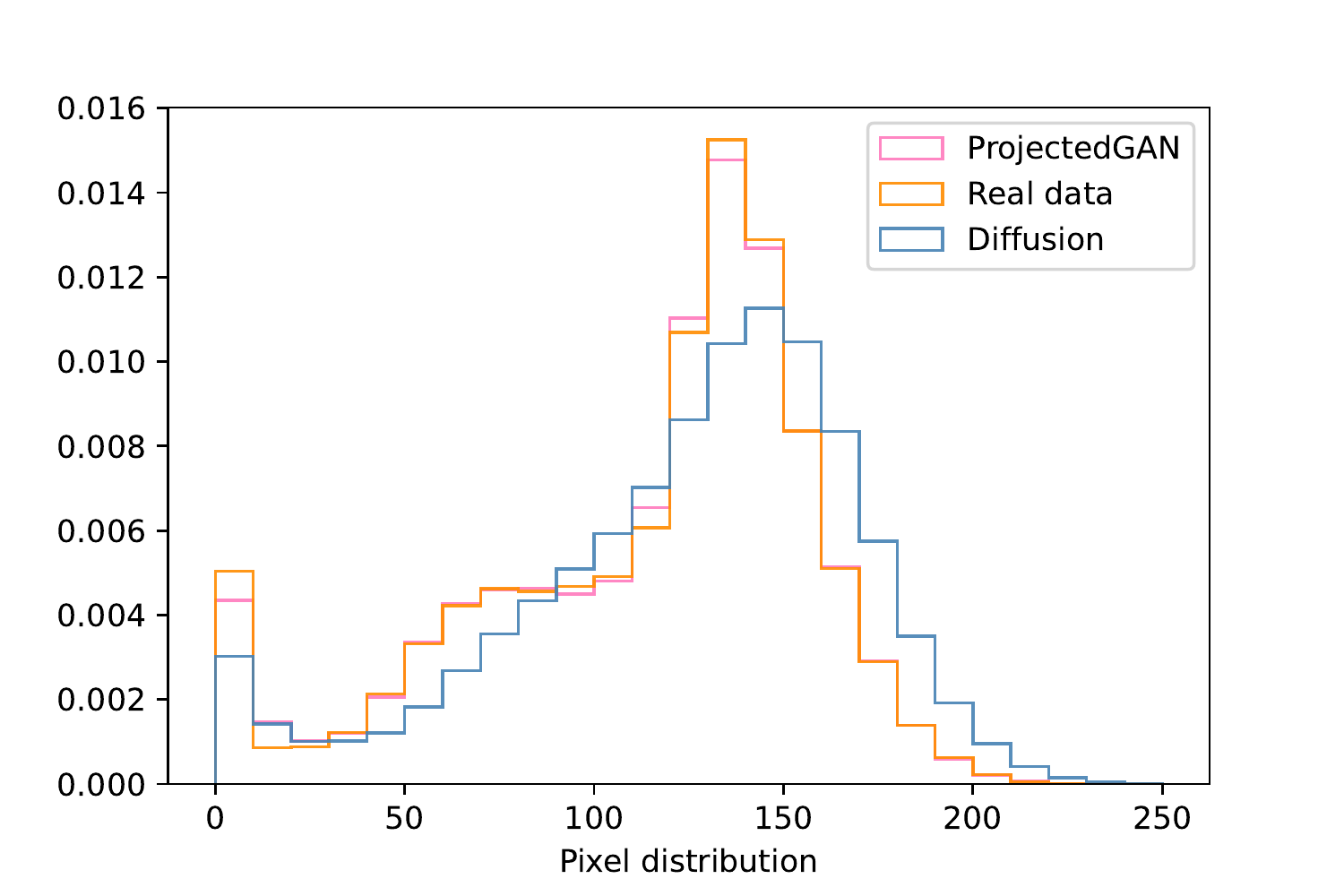}
  \caption{
       Pixel intensities of diffusion model samples and real data do not match together, real data has  heavier tails in the left side and diffusion samples have heavier tails in the right side (cutoff of $150$), while ProjectedGAN and real data match. This could explain the difference in FID we observe between the samples of the two models.}
  \label{fig:histogram}
  \end{center}
\end{figure*}
In our experiments, we measure strikingly distinct FID values for the samples generated by the ProjectedGAN and diffusion-based ADM. This does not correspond to the fine and coarse scale details quality as inspected visually - both models are good in capturing such important details as coronal loops inside and outside the sun disk as well as spherical disk shape. To understand the reasons behind FID differences, we were further analyzing the pixel intensity distribution, comparing distributions underlying real solar images and the generated samples from ProjectedGAN and ADM (Fig.~\ref{fig:histogram}). 

We observe that  the pixel distribution (with values ranging from 0 to 255)  generated by diffusion (ADM) deviates from the real one: the real data has heaver left tail (cutoff 150) while ADM has a heavier right tail, resulting in mean pixel value of $113$ for real data and $127$ for ADM. Opposed to that, ProjectedGAN matches not only the pixel mean ($113$), but also the left and right tails. Thus, the difference in FID is well reflected in different degree of matching real data distribution between ProjectedGAN and ADM. Here, the conclusion is in line with previous observations stating that FID alone cannot serve as reliable measure of generated samples quality - as samples of similar quality may have strongly different FID scores, clearly evident in our observation.   

\subsection{Hyperparameters}

Following~\cite{dhariwal2021diffusion}, we use 128 base channels, 2 residual blocks per resolution, attention layers in resolution 16 and 8 with 4 heads, and a linear noise schedule. For training, we use a learning rate of $0.0001$, a batch size of $64$, use an EMA rate of $0.9999$ and train for 100K steps. For evaluation, we select the model checkpoint with the best FID.

\begin{figure}[!h]
\centering
\begin{subfigure}{0.48\textwidth}
  \centering
  \includegraphics[width=\linewidth]{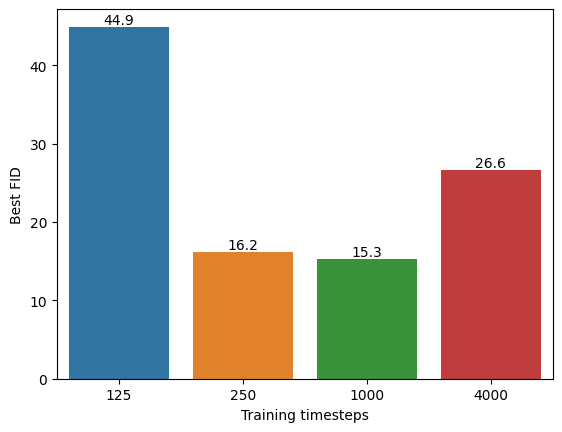}
  \caption{}
  \label{fig:diffusion_hypers_1}
\end{subfigure}
\hfill
\begin{subfigure}{0.48\textwidth}
  \centering
  \includegraphics[width=\linewidth]{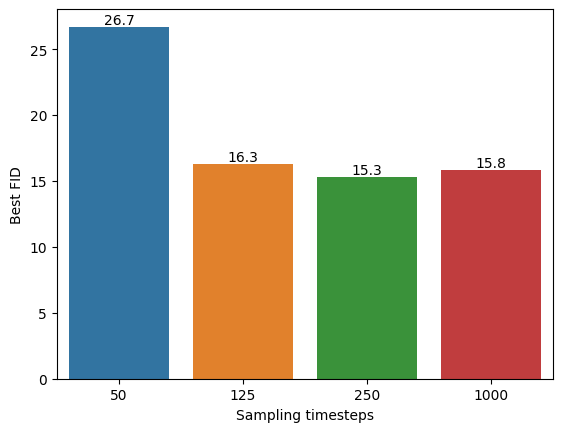}
  \caption{}
  \label{fig:diffusion_hypers_2}
\end{subfigure}
\caption{Effect of training  and sampling  denoising timesteps (\ref{fig:diffusion_hypers_1}) on diffusion model performance (FID). In \ref{fig:diffusion_hypers_1}, we show the best obtained FID with models trained with different training denoising timesteps. In \ref{fig:diffusion_hypers_1}, we show the best FID obtained when varying the sampling timesteps with DDPM for a fixed model trained with 1000 denoising timesteps.}
\label{fig:diffusion_hypers}
\end{figure}

We trained diffusion models with different training/sampling timesteps and with/without regularization. In Fig.~\ref{fig:diffusion_hypers_1}, we show the best FID obtained with number of training timesteps with models trained up to 100K iterations. In ~\ref{fig:diffusion_hypers_2}, we show the best FID obtained when varying the sampling timesteps for a fixed model trained with 1000 denoising timesteps.  After observing initially that FID starts to increase after reaching 60K iterations, we attempted to regularize the model or reduce capacity (using 1 residual block per resolution, attention layers with 2 heads,  and 32 base channels), results are shown in Tab.~\ref{tab:diffusion_hypers_reg}. 

Overall, the best combination we found is 1000 training timesteps, 250 sampling timesteps with random horizontal flipping. We use the best model  in our comparisons with ProjectedGAN models.

\begin{table}[!h]
\begin{center}
\begin{tabular}{lr}
\toprule
                                          Model &  FID \\
\midrule
Random horizontal flip + Reduced model capacity & 43.3 \\
                              No regularization & 28.1 \\
                                    Dropout=0.3 & 25.4 \\
                         Random Horizontal flip & 15.3 \\
\bottomrule
\end{tabular}

\caption{Effect of regularization and model size on diffusion model performance}
\label{tab:diffusion_hypers_reg}
\end{center}
\end{table}

\section{Additional details on evaluation}

We evaluate the models using FID, rFID, KID, CLIP-RN50 based FID, precision, and recall.
We randomly sample 50K images from each model.
We provide detailed evaluation metrics in Tab.~\ref{table:metrics} and Tab.~\ref{table:metrics-patch}. In Tab.~\ref{table:metrics-corr}, we measure the aggreement between the different metrics using Spearman rank correlation.

First, we note that from a model selection perspective, the best model according to each metric will lead to a different choice, which makes it hard to automatize model selection, hence human assessment is still very helpful, especially in domain specific datasets like solar data. We note however that our best ProjGAN model (baseline) achieves systematically  either first or second rank under all the metrics we  consider in Tab~.\ref{table:metrics} and Tab.~\ref{table:metrics-patch}. On the other hand,
our best diffusion model have a poor performance in all metrics compared to our ProjectedGAN-based models, despite being the best among assessed image quality on fine and coarse scale.


\begin{table}[h]
\scriptsize
\begin{tabular}{lrrrrrr}
\toprule
                                          Model &    FID $\downarrow$&    rFID ($\times 10^3$)$\downarrow$ &    KID($\times 10^3$)$\downarrow$ &  CLIP-FID ($\times 10^3$)$\downarrow$&  Prec.$\uparrow$ &  Rec$.\uparrow$ \\
\midrule
                        ProjectedGAN (Baseline) &   \textbf{2.37} &   10.79 &   \textbf{0.74} &      12.10 &       0.60 &    \textbf{0.84} \\
                             EfficientNet-Lite3 &   3.80 &   13.08 &   1.61 &      20.42 &       0.54 &    0.71 \\
                             EfficientNet-Lite2 &   4.07 &   13.17 &   0.99 &      13.43 &       0.58 &    0.75 \\
                              Augmentations off &   4.19 &   17.34 &   1.62 &      \textbf{10.81} &       \textbf{0.66} &    0.51 \\
                            Discriminator 1,2,3 &   6.45 &   19.25 &   2.50 &      20.48 &       0.65 &    0.29 \\
                          Trainable Projections &   7.22 &   15.12 &   3.88 &      31.89 &       0.48 &    0.57 \\
                             EfficientNet-Lite1 &   7.42 &   18.14 &   4.31 &      24.91 &       0.63 &    0.47 \\
                             Discriminator 1,2, &   7.43 &   24.99 &   3.15 &      26.35 &       0.54 &    0.33 \\
                          No Cross Scale Mixing &   9.60 &   28.75 &   2.89 &  22.91 &       0.60 &    0.15 \\
                                Discriminator 1 &  10.69 &   22.40 &   6.15 &      37.54 &       0.47 &    0.39 \\
                  No Cross Scale/Channel Mixing &  10.99 &   32.29 &   4.74 &      27.56 &       0.62 &    0.16 \\
                                Diffusion (ADM) &  15.27 &  140.63 &  15.59 &     111.25 &       0.43 &    0.63 \\
                         Random feature network &  17.72 &   \textbf{9.01} &  16.44 &     267.15 &       0.15 &    0.57 \\
                       Unfreeze feature network & 171.56 & 3366.57 & 199.23 &    2440.33 &       0.01 &    0.00 \\
Randomly initialized feature network (unfrozen) & 252.43 & 5119.63 & 299.56 &    3111.04 &       0.00 &    0.00 \\
 Unfreeze feature network+Trainable Projections & 328.04 & 7221.13 & 405.74 &    4784.95 &       0.00 &    0.00 \\
\bottomrule
\end{tabular}
\caption{Detailed performance metrics of the models.Best model on each metric is highlighted in \textbf{bold}.}
\label{table:metrics}
\end{table}

We further investigated whether pre-training domain specific models and using them for evaluation can agree better with human assessment of solar data image generation. We pre-trained a masked-autoencoder~\cite{he2022masked} (MAE) and VicReg~\cite{bardes2021vicreg} on solar data, and used them to compute Fréchet Distance (FD) between real and generated samples of the different models, and also compare them with  MAE and VicReg models pre-trained on ImageNet.
For MAE pre-training on solar data, we used a Vit-B/16 model with 75\%  masking ratio and we pre-trained the model for 1600 epochs. For VicReg on solar data, we pre-trained a ResNet50 for 1000 epochs.
For MAE and VicReg pre-trained on ImageNet data, we used openly available checkpoints for B/16\footnote{\url{https://github.com/facebookresearch/mae\#fine-tuning-with-pre-trained-checkpoints}} and ResNet50 models\footnote{\url{https://github.com/facebookresearch/vicreg\#pretrained-models-on-pytorch-hub}} respectively.

In Tab.~\ref{table:metrics_ext}, We observe that with an MAE pre-trained on solar data, our best diffusion model is ranked second best (and better than ProjGAN baseline), while it is ranked poorly using both original FID and with MAE pre-trained on ImageNet. With VicReg, we observe a different outcome, there the best diffusion model is ranked poorly both with VicReg pre-trained on ImageNet and on solar data. We also observe that the ProjGAN with a randomly initialized feature network is ranked first when using MAE or VicReg pre-trained on solar data, although it has poorer sample quality than the best ProjGAN and diffusion models (human assessment).

As a conclusion, model ranking can be significantly impacted by the pre-training data used to compute Frechet distances (in our case, ImageNet vs solar data).   This finding emphasizes the need for further investigation into the impact of pre-training data (used to train models used for evaluation) on model selection of existing generative models. 

\begin{table}[h]
\scriptsize

\begin{tabular}{lrrrrr}
\toprule
                                          Model &    FID &  MAE-IN-FD &  MAE-SOL-FD &  VIC-IN-FD &  VIC-SOL-FD \\
\midrule
                        ProjectedGAN (Baseline) &   \textbf{2.37} &          \textbf{8.44} &          29.49 &             \textbf{3.29} &              {\underline{4.99}} \\
                             EfficientNet-Lite3 &   {\underline{3.80}} &         12.44 &          29.87 &             4.88 &              5.80 \\
                             EfficientNet-Lite2 &   4.07 &         11.14 &          29.88 &             {\underline{4.45}} &              5.52 \\
                              Augmentations off &   4.19 &          {\underline{9.55}} &          29.07 &             5.03 &              5.59 \\
                            Discriminator 1,2,3 &   6.45 &         14.10 &          29.77 &             7.84 &              6.68 \\
                          Trainable Projections &   7.22 &         16.09 &          30.72 &             8.42 &              6.20 \\
                             EfficientNet-Lite1 &   7.42 &         18.53 &          30.11 &             8.33 &              6.53 \\
                              Discriminator 1,2 &   7.43 &         20.29 &          31.50 &             8.40 &              7.08 \\
                          No Cross Scale Mixing &   9.60 &         18.79 &          33.27 &            10.21 &              8.46 \\
                                Discriminator 1 &  10.69 &         27.06 &          31.71 &            10.40 &              7.76 \\
                  No Cross Scale/Channel Mixing &  10.99 &         21.73 &          32.09 &            10.04 &              8.90 \\
                                      Diffusion &  15.27 &         53.80 &          \underline{28.92} &            19.55 &             12.10 \\
                         Random feature network &  17.72 &         27.16 &          \textbf{28.67} &            23.40 &              \textbf{4.59} \\
                      Unfreeze feature network & 171.56 &       1141.76 &          66.04 &           365.75 &            494.73 \\
 Random feature network (unfrozen) & 252.43 &       1727.87 &          78.49 &           565.27 &           1068.64 \\
Unfreeze feature network + Train projs & 328.04 &       4162.40 &         148.02 &           555.81 &            140.88 \\
\bottomrule
\end{tabular}
\caption{Additional performance metrics of the models based on pre-trained MAE and VicReg (noted VIC) models to compare between Fréchet Distances based on ImageNet (noted IN) and on solar data (noted SOL).
Best model on each metric is highlighted in \textbf{bold}, while second best is \underline{underlined}.
}
\label{table:metrics_ext}
\end{table}

\begin{table}[h]
\small
\begin{center}
\begin{tabular}{lrrrr}
\toprule
                                          Model &    FID &  FID-p64 &  FID-p128 &  FID-p256 \\
\midrule
                        ProjectedGAN (Baseline) &   \textbf{2.37} &     1.01 &      \textbf{0.84} &      0.97 \\
                             EfficientNet-Lite3 &   3.80 &     \textbf{0.96} &      0.99 &      1.24 \\
                             EfficientNet-Lite2 &   4.07 &     1.05 &      0.90 &      \textbf{0.96} \\
                              Augmentations off &   4.19 &     3.68 &      3.55 &      1.62 \\
                            Discriminator 1,2,3 &   6.45 &     1.07 &      1.18 &      1.79 \\
                          Trainable Projections &   7.22 &     2.37 &      2.15 &      2.60 \\
                             EfficientNet-Lite1 &   7.42 &     1.28 &      1.59 &      2.08 \\
                             Discriminator 1,2, &   7.43 &     1.37 &      1.73 &      2.26 \\
                  No Cross Scale Mixing &   9.60 &     1.18 &      1.51 &      2.55 \\
                                Discriminator 1 &  10.69 &     2.54 &      2.86 &      4.00 \\
                  No Cross Scale/Channel Mixing &  10.99 &     1.65 &      1.53 &      2.85 \\
                                      Diffusion (ADM) &  15.27 &     8.99 &      7.92 &     11.83 \\
                         Random feature network &  17.72 &    21.80 &     36.15 &     55.84 \\
                       Unfreeze feature network & 171.56 &   134.92 &    128.03 &    162.78 \\
Random feature network (unfrozen) & 252.43 &   191.38 &    175.69 &    187.80 \\
 Unfreeze feature network+Trainable Projections & 328.04 &   414.97 &    441.43 &    506.61 \\
\bottomrule
\end{tabular}
\caption{Results on FID and Patch-based FID, where FID is computed on patches, e.g., FID-p64 is computed by using patches of size 64x64 extracted randomly from images.}
\label{table:metrics-patch}
\end{center}
\end{table}

\begin{table}[h]
\scriptsize
\begin{center}
\begin{tabular}{lrrrrrrrrrr}
\toprule
 &   FID &  rFID &   KID &  CLIP-RN50 &  Prec. &  Rec. &  MAE-IN-FD &  MAE-SOL-FD &  VIC-IN-FD &  VIC-SOL-FD \\
               &       &       &       &            &            &         &               &                &                  &                   \\
\midrule
FID              &  1.00 &  0.75 &  0.97 &       0.94 &      -0.74 &   -0.71 &          \textbf{0.98} &           0.53 &             \textbf{0.98} &              0.78 \\
rFID             &  0.75 &  1.00 &  0.68 &       0.64 &      -0.44 &   -0.83 &          0.75 &           0.73 &             0.71 &              \textbf{0.98} \\
KID              &  \textbf{0.97} &  0.68 &  1.00 &       \textbf{0.97} &      -0.76 &   -0.64 &          0.96 &           0.48 &             0.96 &              0.72 \\
CLIP-RN50        &  0.94 &  0.64 &  \textbf{0.97} &       1.00 &      -0.86 &   -0.58 &          0.96 &           0.51 &             0.95 &              0.69 \\
Prec.        & -0.74 & -0.44 & -0.76 &      \textbf{-0.86} &       1.00 &    0.31 &         -0.81 &          -0.41 &            -0.79 &             -0.48 \\
Rec.          & -0.71 & \textbf{-0.83} & -0.64 &      -0.58 &       0.31 &    1.00 &         -0.66 &          -0.81 &            -0.69 &             \textbf{-0.83} \\
MAE-IN-FD     &  \textbf{0.98} &  0.75 &  0.96 &       0.96 &      -0.81 &   -0.66 &          1.00 &           0.54 &             0.96 &              0.79 \\
MAE-SOL-FD    &  0.53 &  0.73 &  0.48 &       0.51 &      -0.41 &   \textbf{-0.81} &          0.54 &           1.00 &             0.53 &              0.72 \\
VIC-IN-FD  &  \textbf{0.98} &  0.71 &  0.96 &       0.95 &      -0.79 &   -0.69 &          0.96 &           0.53 &             1.00 &              0.76 \\
VIC-SOL-FD &  0.78 &  \textbf{0.98} &  0.72 &       0.69 &      -0.48 &   -0.83 &          0.79 &           0.72 &             0.76 &              1.00 \\
\bottomrule
\end{tabular}

\caption{We measure agreement between performance metrics using Spearman rank correlation ($\rho$). For each metric (rows), we highlight in \textbf{bold} the metric with highest correlation.
}
\label{table:metrics-corr}
\end{center}
\end{table}




\end{appendix}

%% file: main.bbl
\begin{thebibliography}{10}\itemsep=-1pt

\bibitem{bardes2021vicreg}
Adrien Bardes, Jean Ponce, and Yann LeCun.
\newblock Vicreg: Variance-invariance-covariance regularization for
  self-supervised learning.
\newblock {\em arXiv preprint arXiv:2105.04906}, 2021.

\bibitem{Barnes2020}
Will {Barnes}, Mark {Cheung}, Monica {Bobra}, Paul {Boerner}, Georgios
  {Chintzoglou}, Drew {Leonard}, Stuart {Mumford}, Nicholas {Padmanabhan},
  Albert {Shih}, Nina {Shirman}, David {Stansby}, and Paul {Wright}.
\newblock {aiapy: A Python Package for Analyzing Solar EUV Image Data from
  AIA}.
\newblock {\em The Journal of Open Source Software}, 5(55):2801, Nov. 2020.

\bibitem{binkowski2018demystifying}
Miko{\l}aj Bi{\'n}kowski, Danica~J Sutherland, Michael Arbel, and Arthur
  Gretton.
\newblock Demystifying mmd gans.
\newblock {\em arXiv preprint arXiv:1801.01401}, 2018.

\bibitem{Bintsi2022}
Kyriaki-Margarita {Bintsi}, Robert {Jarolim}, Benoit {Tremblay}, Miraflor
  {Santos}, Anna {Jungbluth}, James~Paul {Mason}, Sairam {Sundaresan}, Angelos
  {Vourlidas}, Cooper {Downs}, Ronald~M. {Caplan}, and Andr{\'e}s {Mu{\~n}oz
  Jaramillo}.
\newblock {SuNeRF: Validation of a 3D Global Reconstruction of the Solar Corona
  Using Simulated EUV Images}.
\newblock {\em arXiv e-prints}, page arXiv:2211.14879, Nov. 2022.

\bibitem{Bobra2015}
M.~G. {Bobra} and S. {Couvidat}.
\newblock {Solar Flare Prediction Using SDO/HMI Vector Magnetic Field Data with
  a Machine-learning Algorithm}.
\newblock {\em apj}, 798(2):135, Jan. 2015.

\bibitem{Bobra2016}
M.~G. {Bobra} and S. {Ilonidis}.
\newblock {Predicting Coronal Mass Ejections Using Machine Learning Methods}.
\newblock {\em apj}, 821(2):127, Apr. 2016.

\bibitem{HelioML2020}
Monica~G. {Bobra} and James~P. {Mason}.
\newblock {\em {Machine Learning, Statistics, and Data Mining for
  Heliophysics}}.
\newblock ., 2020.

\bibitem{chifu20213d}
Iulia Chifu and Ricardo Gafeira.
\newblock 3d solar coronal loop reconstructions with machine learning.
\newblock {\em The Astrophysical Journal Letters}, 910(1):L10, 2021.

\bibitem{chong2020effectively}
Min~Jin Chong and David Forsyth.
\newblock Effectively unbiased fid and inception score and where to find them.
\newblock In {\em Proceedings of the IEEE/CVF conference on computer vision and
  pattern recognition}, pages 6070--6079, 2020.

\bibitem{deng2009imagenet}
Jia Deng, Wei Dong, Richard Socher, Li-Jia Li, Kai Li, and Li Fei-Fei.
\newblock Imagenet: A large-scale hierarchical image database.
\newblock In {\em 2009 IEEE conference on computer vision and pattern
  recognition}, pages 248--255. Ieee, 2009.

\bibitem{dhariwal2021diffusion}
Prafulla Dhariwal and Alexander Nichol.
\newblock Diffusion models beat gans on image synthesis.
\newblock {\em Advances in Neural Information Processing Systems},
  34:8780--8794, 2021.

\bibitem{esser2021taming}
Patrick Esser, Robin Rombach, and Bjorn Ommer.
\newblock Taming transformers for high-resolution image synthesis.
\newblock In {\em Proceedings of the IEEE/CVF conference on computer vision and
  pattern recognition}, pages 12873--12883, 2021.

\bibitem{Galvez2019}
Richard Galvez, David~F Fouhey, Meng Jin, Alexandre Szenicer, Andr{\'e}s
  Mu{\~n}oz-Jaramillo, Mark~CM Cheung, Paul~J Wright, Monica~G Bobra, Yang Liu,
  James Mason, et~al.
\newblock A machine-learning data set prepared from the nasa solar dynamics
  observatory mission.
\newblock {\em The Astrophysical Journal Supplement Series}, 242(1):7, 2019.

\bibitem{goodfellow2014generative}
Ian Goodfellow, Jean Pouget-Abadie, Mehdi Mirza, Bing Xu, David Warde-Farley,
  Sherjil Ozair, Aaron Courville, and Yoshua Bengio.
\newblock Generative adversarial nets.
\newblock {\em Advances in Neural Information Processing Systems}, 27, 2014.

\bibitem{harkonen2020ganspace}
Erik H{\"a}rk{\"o}nen, Aaron Hertzmann, Jaakko Lehtinen, and Sylvain Paris.
\newblock Ganspace: Discovering interpretable gan controls.
\newblock {\em Advances in Neural Information Processing Systems},
  33:9841--9850, 2020.

\bibitem{he2022masked}
Kaiming He, Xinlei Chen, Saining Xie, Yanghao Li, Piotr Doll{\'a}r, and Ross
  Girshick.
\newblock Masked autoencoders are scalable vision learners.
\newblock In {\em Proceedings of the IEEE/CVF Conference on Computer Vision and
  Pattern Recognition}, pages 16000--16009, 2022.

\bibitem{heusel2017gans}
Martin Heusel, Hubert Ramsauer, Thomas Unterthiner, Bernhard Nessler, and Sepp
  Hochreiter.
\newblock Gans trained by a two time-scale update rule converge to a local nash
  equilibrium.
\newblock {\em Advances in neural information processing systems}, 30, 2017.

\bibitem{ho2020denoising}
Jonathan Ho, Ajay Jain, and Pieter Abbeel.
\newblock Denoising diffusion probabilistic models.
\newblock {\em Advances in Neural Information Processing Systems},
  33:6840--6851, 2020.

\bibitem{Jeong2022}
Hyun-Jin {Jeong}, Yong-Jae {Moon}, Eunsu {Park}, Harim {Lee}, and Ji-Hye
  {Baek}.
\newblock {Improved AI-generated Solar Farside Magnetograms by STEREO and SDO
  Data Sets and Their Release}.
\newblock {\em apjs}, 262(2):50, Oct. 2022.

\bibitem{Jonas2018}
Eric {Jonas}, Monica {Bobra}, Vaishaal {Shankar}, J. {Todd Hoeksema}, and
  Benjamin {Recht}.
\newblock {Flare Prediction Using Photospheric and Coronal Image Data}.
\newblock {\em solphys}, 293(3):48, Mar. 2018.

\bibitem{karras2020training}
Tero Karras, Miika Aittala, Janne Hellsten, Samuli Laine, Jaakko Lehtinen, and
  Timo Aila.
\newblock Training generative adversarial networks with limited data.
\newblock {\em Advances in Neural Information Processing Systems},
  33:12104--12114, 2020.

\bibitem{karras2021alias}
Tero Karras, Miika Aittala, Samuli Laine, Erik H{\"a}rk{\"o}nen, Janne
  Hellsten, Jaakko Lehtinen, and Timo Aila.
\newblock Alias-free generative adversarial networks.
\newblock {\em Advances in Neural Information Processing Systems}, 34:852--863,
  2021.

\bibitem{karras2019style}
Tero Karras, Samuli Laine, and Timo Aila.
\newblock A style-based generator architecture for generative adversarial
  networks.
\newblock In {\em Proceedings of the IEEE/CVF conference on computer vision and
  pattern recognition}, pages 4401--4410, 2019.

\bibitem{karras2020analyzing}
Tero Karras, Samuli Laine, Miika Aittala, Janne Hellsten, Jaakko Lehtinen, and
  Timo Aila.
\newblock Analyzing and improving the image quality of stylegan.
\newblock In {\em Proceedings of the IEEE/CVF conference on computer vision and
  pattern recognition}, pages 8110--8119, 2020.

\bibitem{Kasapis2022}
Spiridon {Kasapis}, Lulu {Zhao}, Yang {Chen}, Xiantong {Wang}, Monica {Bobra},
  and Tamas {Gombosi}.
\newblock {Interpretable Machine Learning to Forecast SEP Events for Solar
  Cycle 23}.
\newblock {\em Space Weather}, 20(2):e2021SW002842, Feb. 2022.

\bibitem{Kim2019}
Taeyoung {Kim}, Eunsu {Park}, Harim {Lee}, Yong-Jae {Moon}, Sung-Ho {Bae}, Daye
  {Lim}, Soojeong {Jang}, Lokwon {Kim}, Il-Hyun {Cho}, Myungjin {Choi}, and
  Kyung-Suk {Cho}.
\newblock {Solar farside magnetograms from deep learning analysis of
  STEREO/EUVI data}.
\newblock {\em Nature Astronomy}, 3:397--400, Mar. 2019.

\bibitem{langley00}
P. Langley.
\newblock Crafting papers on machine learning.
\newblock In Pat Langley, editor, {\em Proceedings of the 17th International
  Conference on Machine Learning (ICML 2000)}, pages 1207--1216, Stanford, CA,
  2000. Morgan Kaufmann.

\bibitem{Lemen2012SoPh}
James~R. {Lemen}, Alan~M. {Title}, David~J. {Akin}, Paul~F. {Boerner},
  Catherine {Chou}, Jerry~F. {Drake}, Dexter~W. {Duncan}, Christopher~G.
  {Edwards}, Frank~M. {Friedlaender}, Gary~F. {Heyman}, Neal~E. {Hurlburt},
  Noah~L. {Katz}, Gary~D. {Kushner}, Michael {Levay}, Russell~W. {Lindgren},
  Dnyanesh~P. {Mathur}, Edward~L. {McFeaters}, Sarah {Mitchell}, Roger~A.
  {Rehse}, Carolus~J. {Schrijver}, Larry~A. {Springer}, Robert~A. {Stern},
  Theodore~D. {Tarbell}, Jean-Pierre {Wuelser}, C.~Jacob {Wolfson}, Carl
  {Yanari}, Jay~A. {Bookbinder}, Peter~N. {Cheimets}, David {Caldwell},
  Edward~E. {Deluca}, Richard {Gates}, Leon {Golub}, Sang {Park}, William~A.
  {Podgorski}, Rock~I. {Bush}, Philip~H. {Scherrer}, Mark~A. {Gummin}, Peter
  {Smith}, Gary {Auker}, Paul {Jerram}, Peter {Pool}, Regina {Soufli}, David~L.
  {Windt}, Sarah {Beardsley}, Matthew {Clapp}, James {Lang}, and Nicholas
  {Waltham}.
\newblock {The Atmospheric Imaging Assembly (AIA) on the Solar Dynamics
  Observatory (SDO)}.
\newblock {\em solphys}, 275(1-2):17--40, Jan. 2012.

\bibitem{liu2022pseudo}
Luping Liu, Yi Ren, Zhijie Lin, and Zhou Zhao.
\newblock Pseudo numerical methods for diffusion models on manifolds.
\newblock {\em arXiv preprint arXiv:2202.09778}, 2022.

\bibitem{meng2022distillation}
Chenlin Meng, Ruiqi Gao, Diederik~P Kingma, Stefano Ermon, Jonathan Ho, and Tim
  Salimans.
\newblock On distillation of guided diffusion models.
\newblock {\em arXiv preprint arXiv:2210.03142}, 2022.

\bibitem{Mercea2023}
Vanessa Mercea, Alin~Razvan Paraschiv, Daniela~Adriana Lacatus, Anca Marginean,
  and Diana Besliu-Ionescu.
\newblock A machine learning enhanced approach for automated sunquake detection
  in acoustic emission maps.
\newblock {\em Solar Physics}, 298(1):4, 2023.

\bibitem{Park2019}
Eunsu {Park}, Yong-Jae {Moon}, Jin-Yi {Lee}, Rok-Soon {Kim}, Harim {Lee}, Daye
  {Lim}, Gyungin {Shin}, and Taeyoung {Kim}.
\newblock {Generation of Solar UV and EUV Images from SDO/HMI Magnetograms by
  Deep Learning}.
\newblock {\em apjl}, 884(1):L23, Oct. 2019.

\bibitem{pesnell2012solar}
W~Dean Pesnell, B.J. Thompson, and PC Chamberlin.
\newblock {\em The solar dynamics observatory (SDO)}.
\newblock Springer, 2012.

\bibitem{ramesh2022hierarchical}
Aditya Ramesh, Prafulla Dhariwal, Alex Nichol, Casey Chu, and Mark Chen.
\newblock Hierarchical text-conditional image generation with clip latents.
\newblock {\em arXiv preprint arXiv:2204.06125}, 2022.

\bibitem{reiss2015}
Martin~A. {Reiss}, Stefan~J. {Hofmeister}, Ruben {De Visscher}, Manuela
  {Temmer}, Astrid~M. {Veronig}, V{\'e}ronique {Delouille}, Benjamin {Mampaey},
  and Helmut {Ahammer}.
\newblock {Improvements on coronal hole detection in SDO/AIA images using
  supervised classification}.
\newblock {\em Journal of Space Weather and Space Climate}, 5:A23, July 2015.

\bibitem{rombach2022high}
Robin Rombach, Andreas Blattmann, Dominik Lorenz, Patrick Esser, and Bj{\"o}rn
  Ommer.
\newblock High-resolution image synthesis with latent diffusion models.
\newblock In {\em Proceedings of the IEEE/CVF Conference on Computer Vision and
  Pattern Recognition}, pages 10684--10695, 2022.

\bibitem{saharia2022photorealistic}
Chitwan Saharia, William Chan, Saurabh Saxena, Lala Li, Jay Whang, Emily
  Denton, Seyed Kamyar~Seyed Ghasemipour, Burcu~Karagol Ayan, S~Sara Mahdavi,
  Rapha~Gontijo Lopes, et~al.
\newblock Photorealistic text-to-image diffusion models with deep language
  understanding.
\newblock {\em arXiv preprint arXiv:2205.11487}, 2022.

\bibitem{salimans2022progressive}
Tim Salimans and Jonathan Ho.
\newblock Progressive distillation for fast sampling of diffusion models.
\newblock {\em arXiv preprint arXiv:2202.00512}, 2022.

\bibitem{sauer2021projected}
Axel Sauer, Kashyap Chitta, Jens M{\"u}ller, and Andreas Geiger.
\newblock Projected gans converge faster.
\newblock {\em Advances in Neural Information Processing Systems},
  34:17480--17492, 2021.

\bibitem{sauer2022stylegan}
Axel Sauer, Katja Schwarz, and Andreas Geiger.
\newblock Stylegan-xl: Scaling stylegan to large diverse datasets.
\newblock In {\em ACM SIGGRAPH 2022 conference proceedings}, pages 1--10, 2022.

\bibitem{sohl2015deep}
Jascha Sohl-Dickstein, Eric Weiss, Niru Maheswaranathan, and Surya Ganguli.
\newblock Deep unsupervised learning using nonequilibrium thermodynamics.
\newblock In {\em International Conference on Machine Learning}, pages
  2256--2265. PMLR, 2015.

\bibitem{song2020denoising}
Jiaming Song, Chenlin Meng, and Stefano Ermon.
\newblock Denoising diffusion implicit models.
\newblock {\em arXiv preprint arXiv:2010.02502}, 2020.

\bibitem{song2019generative}
Yang Song and Stefano Ermon.
\newblock Generative modeling by estimating gradients of the data distribution.
\newblock {\em Advances in neural information processing systems}, 32, 2019.

\bibitem{Sun2022}
Wenqing {Sun}, Long {Xu}, Suli {Ma}, Yihua {Yan}, Tie {Liu}, and Weiqiang
  {Zhang}.
\newblock {A Dynamic Deep-learning Model for Generating a Magnetogram Sequence
  from an SDO/AIA EUV Image Sequence}.
\newblock {\em apjs}, 262(2):45, Oct. 2022.

\bibitem{szegedy2016rethinking}
Christian Szegedy, Vincent Vanhoucke, Sergey Ioffe, Jon Shlens, and Zbigniew
  Wojna.
\newblock Rethinking the inception architecture for computer vision.
\newblock In {\em Proceedings of the IEEE conference on computer vision and
  pattern recognition}, pages 2818--2826, 2016.

\bibitem{theis2015note}
Lucas Theis, A{\"a}ron van~den Oord, and Matthias Bethge.
\newblock A note on the evaluation of generative models.
\newblock {\em arXiv preprint arXiv:1511.01844}, 2015.

\bibitem{xiao2021tackling}
Zhisheng Xiao, Karsten Kreis, and Arash Vahdat.
\newblock Tackling the generative learning trilemma with denoising diffusion
  gans.
\newblock {\em arXiv preprint arXiv:2112.07804}, 2021.

\bibitem{Yu2021}
Xuexin {Yu}, Long {Xu}, and Yihua {Yan}.
\newblock {Image Desaturation for SDO/AIA Using Deep Learning}.
\newblock {\em solphys}, 296(3):56, Mar. 2021.

\bibitem{zhao2020differentiable}
Shengyu Zhao, Zhijian Liu, Ji Lin, Jun-Yan Zhu, and Song Han.
\newblock Differentiable augmentation for data-efficient gan training.
\newblock {\em Advances in Neural Information Processing Systems},
  33:7559--7570, 2020.

\end{thebibliography}
